\documentclass[sn-mathphys]{sn-jnl}
\usepackage{cleveref}

\jyear{2022}%

\newcommand{\Jfunc}{\Psi}                    
\newcommand{\nstate}{n}

\newif\ifcomplete
\newif\ifdraft

\newcommand{\sfrac}[2]{\mbox{\footnotesize$\displaystyle\frac{#1}{#2}$}} 

\newcommand{\Hobs}{\mathcal{H}}
\newcommand{\HH}{\mathbf H}
\newcommand{\Nobs}{ {{\rm N}_{\rm obs}} }

\newcommand{\danB}{ \mathbf{B} }

\newcommand{\Id}{ \mathbf{I} }
\newcommand{\danR}{ \mathbf{R} }

\newcommand{\danQ}{\mathbf{Q}}

\newcommand{\Model}{\mathcal{M}}

\newcommand{\Mtlm}{\mathbf{M}}

\newcommand{\danx}[1][]{%
   \ifthenelse{ \equal{#1}{} }
      {\mathbf{x}}
      {\mathbf{x}^{[#1]}}}
\newcommand{\xb}[1][]{%
   \ifthenelse{ \equal{#1}{} }
      {\mathbf{x}^{\rm b}}
      {\mathbf{x}^{{\rm b}[#1]}}}
\newcommand{\xa}[1][]{%
   \ifthenelse{ \equal{#1}{} }
      {\mathbf{x}^{\rm a}}
      {\mathbf{x}^{{\rm a}[#1]}}}
\newcommand{\xf}[1][]{%
   \ifthenelse{ \equal{#1}{} }
      {\mathbf{y}}
      {\mathbf{y}^{[#1]}}}
\newcommand{\dany}[1][]{%
   \ifthenelse{ \equal{#1}{} }
      {\mathbf{y}}
      {\mathbf{y}^{[#1]}}}

\newcommand{\bigdot}[1]{\accentset{\mbox{\large\bfseries .}}{#1}}
\newcommand{\dx}[1][]{%
   \ifthenelse{ \equal{#1}{} }
      {\bigdot{\mathbf{x}}}
      {\bigdot{\mathbf{x}}^{[#1]}}}
\newcommand{\dxb}[1][]{%
   \ifthenelse{ \equal{#1}{} }
      {\bigdot{\mathbf{x}}^{\rm b}}
      {\bigdot{\mathbf{x}}^{{\rm b}[#1]}}}
\newcommand{\dxa}[1][]{%
   \ifthenelse{ \equal{#1}{} }
      {\bigdot{\mathbf{x}}^{\rm a}}
      {\bigdot{\mathbf{x}}^{{\rm a}[#1]}}}

\newcommand{\xt}{ \mathbf{x}^{\rm true} }

\newcommand{\hofx}[1][]{%
   \ifthenelse{ \equal{#1}{} }
      {\mathbf{z}}
      {\mathbf{z}^{[#1]}}}
\newcommand{\hofxb}[1][]{%
   \ifthenelse{ \equal{#1}{} }
      {\mathbf{z}^{\rm b}}
      {\mathbf{z}^{{\rm b}[#1]}}}
\newcommand{\hofxa}[1][]{%
   \ifthenelse{ \equal{#1}{} }
      {\mathbf{z}^{\rm a}}
      {\mathbf{z}^{{\rm a}[#1]}}}
\newcommand{\dhofxb}[1][]{%
   \ifthenelse{ \equal{#1}{} }
      {\bigdot{\mathbf{z}}^{\rm b}}
      {\bigdot{\mathbf{z}}^{{\rm b}[#1]}}}
\newcommand{\dhofxa}[1][]{%
   \ifthenelse{ \equal{#1}{} }
      {\bigdot{\mathbf{z}}^{\rm a}}
      {\bigdot{\mathbf{z}}^{{\rm a}[#1]}}}
\renewcommand{\d}[1][]{%
   \ifthenelse{ \equal{#1}{} }
      {\mathbf{d}}
      {\mathbf{d}^{[#1]}}}

\newcommand{\errb}[1][]{%
   \ifthenelse{ \equal{#1}{} }
      {\varepsilon^{\rm b}}
      {\varepsilon^{{\rm b}[#1]}}}
\newcommand{\erra}[1][]{%
   \ifthenelse{ \equal{#1}{} }
      {\varepsilon^{\rm a}}
      {\varepsilon^{{\rm a}[#1]}}}
\newcommand{\erro}[1][]{%
   \ifthenelse{ \equal{#1}{} }
      {\varepsilon^{\rm obs}}
      {\varepsilon^{{\rm obs}[#1]}}}

\newcommand{\errm}[1][]{%
   \ifthenelse{ \equal{#1}{} }
      {\varepsilon^{\rm mod}}
      {\varepsilon^{{\rm mod}[#1]}}}
\newcommand{\wb}[1][]{%
   \ifthenelse{ \equal{#1}{} }
      {w^{\rm b}}
      {w^{{\rm b}[#1]}}}
\newcommand{\wa}[1][]{%
   \ifthenelse{ \equal{#1}{} }
      {w^{\rm a}}
      {w^{{\rm a}[#1]}}}

\renewcommand{\Re}{\mathds{R}}

\renewcommand{\H}{\mathcal{H}}
\newcommand{\from}{\sim}

\newcommand{\norm}[1]{\bigl\Vert #1 \bigr\Vert}

\newcommand{\llam}{ {\boldsymbol\lambda} }

\newcommand{\Nstate}{{\rm N}_{\rm state}}

\newcommand{\Madjt}{\mathbf{M}^T}

\newcommand{\nnmadjt}{\mathbf{N}^{T}}
\newcommand{\nnm}{\mathcal{N}}
\newcommand{\nnmtlm}{\mathbf{N}}
\newcommand{\loss}[1]{\mathcal{L}_\text{#1}}
\def\!#1{\mathcal{#1}}
\def\*#1{\boldsymbol{\mathbf{#1}}}
\def\|#1{\textnormal{#1}}
\def\##1{\mathfrak{#1}}
\renewcommand{\Re}{\mathbb{R}}
\newcommand{\Nsub}[1]{{\rm N}_{\rm #1}}
\newcommand{\uin}{{\mathbf{u}}_{\rm in}}
\newcommand{\uout}{{\mathbf{u}}_{\rm out}}
\newcommand{\zout}{{\mathbf{z}}_{\rm out}}

\def\norm#1{\left\lVert#1\right\rVert}

\DeclareMathOperator*{\argmin}{arg\,min}

\newcommand{\dane}[1][]{%
   \ifthenelse{ \equal{#1}{} }
      {\mathbf{e}}
      {\mathbf{e}^{[#1]}}}

\newcommand{\mycell}[1]{%
  \begin{tabular}[t]{@{}l@{}} #1 \end{tabular}}

\newcommand*{\coloneqq}{\mathrel{\vcenter{\baselineskip0.5ex \lineskiplimit0pt
                     \hbox{\scriptsize.}\hbox{\scriptsize.}}}%
                     =}

\usepackage{tabularx}

\newcommand{\myheadertitle}{Improved ML surrogates for 4D-Var}
\newcommand{\mytitle}{Improving Machine Learning Surrogate Models for Variational Data Assimilation Using Derivative Information}

\begin{document}

\title[\myheadertitle]{\mytitle}

\author*[1]{\fnm{Austin} \sur{Chennault} \email{achennault@vt.edu} }

\author[1,2]{\fnm{Andrey A.} \sur{Popov} \email{apopov@vt.edu} }

\author[1]{\fnm{Amit N.} \sur{Subrahmanya} \email{amitns@vt.edu} }

\author[1]{\fnm{Rachel} \sur{Cooper} \email{cyuas@vt.edu} }

\author[1]{\fnm{Ali Haisam Muhammad} \sur{Rafid} \email{haisamrafid@vt.edu} }

\author[1]{\fnm{Anuj} \sur{Karpatne} \email{karpatne@vt.edu} }

\author[1]{\fnm{Adrian} \sur{Sandu} \email{sandu@cs.vt.edu} }

\newcommand{\sandu}[1]{{\color{red}$\langle$#1$\rangle$}}

\affil*[1]{\orgdiv{Department of Computer Science}, \orgname{Virginia Tech}, \orgaddress{\street{620 Drillfield Dr.}, \city{Blacksburg}, \postcode{24061}, \state{Virginia}, \country{USA}}}

\affil[2]{\orgdiv{Oden Institute for Computational Engineering and Sciences}, \orgname{University of Texas at Austin}, \orgaddress{\street{201 E. 24th Street}, \city{Austin}, \postcode{78712}, \state{Texas}, \country{USA}}}

\abstract{Data assimilation is the process of fusing information from imperfect computer simulations with noisy, sparse measurements of reality to obtain improved estimates of the state or parameters of a dynamical system of interest.  The data assimilation procedures used in many geoscience applications, such as numerical weather forecasting, are variants of the our-dimensional variational (4D-Var) algorithm. The cost of solving the underlying 4D-Var optimization problem is dominated by the cost of repeated forward and adjoint model runs.  This motivates substituting the evaluations of the physical model and its adjoint by fast, approximate surrogate models. Neural networks offer a promising approach for the data-driven creation of surrogate models. The accuracy of the surrogate 4D-Var solution depends on the accuracy with each the surrogate captures both the  forward and the adjoint model dynamics.  We formulate and analyze several approaches to incorporate adjoint  information into the construction of neural network surrogates.  The resulting networks are tested on unseen data and in a sequential data assimilation problem using the Lorenz-63 system. Surrogates constructed using adjoint information demonstrate superior performance on the 4D-Var data assimilation problem compared to a standard neural network surrogate that uses only forward dynamics information.}

\keywords{Data assimilation, Machine learning, Optimization, Inverse problems}

\pacs[MSC Classification]{34A55, 68T07, 90C30, 65L09}

\maketitle

\section{Introduction}
\label{sec:intro}

\footnotetext{This work was supported by NSF through grants CDS\&E-MSS-1953113 and CCF-1613905, by DOE through grant ASCR DE-SC0021313, and by the Computational Science Laboratory at Virginia Tech.}

Many areas in science and engineering rely on complex computational models for the simulation of physical systems. A generic model of the form $[\dany,\danx] = \Model(\theta)$, depends on a set of parameters $\theta$, to produce an output $[\dany,\danx]$ that generally represents physical quantities. Here $\danx$ are hidden state variables, while $\dany$ are observable variables. For example, in a numerical weather prediction simulation, $\theta$ can represent the initial conditions, $\danx$ the atmospheric flow vorticity field, and $\dany$ the wind velocities at particular sensor locations. The inverse problem consists of using noisy measurements of the observable quantities $\dany$, together with the model operator $\Model$, to obtain improved estimates of the parameter value $\theta$. The model operator of interest and its corresponding adjoint may be expensive to evaluate. In our numerical weather prediction example $\Model$ is a nonlinear solution operator that maps the initial conditions to the solutions of the underlying partial differential equations, computed at upwards of $10^{12}$ space-time mesh points \cite{ECMWFIFSP3, vanLeeuwen2015}. Solution of an inverse problem may require thousands of forward model evaluations in the statistical setting or several hundred in the variational setting \cite{Frangos2010, Sandu2019}, with solution cost depending primarily on the expense of the forward model $\Model$. Inexpensive surrogate models can then be used in place of the high fidelity model operator $\Model$ for fast, approximate inversion. 

In this paper we focus on data assimilation, i.e., inverse problems where the underlying models $\Model$ represent time-evolving dynamical systems. Surrogate models have a long history within data assimilation. Previously studied approaches have included proper orthogonal decomposition (POD) and variants \cite{Sandu_2015_POD-inverse-problems}, and various neural network approaches \cite{Farchi2020, Panda2020, DINO, Willard2020, maulik2021efficient}. 

This work proposes the construction of specialized neural-network surrogate models for accelerating computation of the four dimensional variational (4D-Var) solution to the data assimilation problems. The 4D-Var approach calculates a maximum aposteriori estimate of the model parameters $\theta$ by solving a constrained optimization problem. The cost function includes the prior information and the mismatch between model predictions and observations, and the constraints are the equations defining the model $\Model$. Our proposed approach is to replace the high fidelity model constraints with the surrogate model equations, thereby considerably reducing the cost of solving the optimization problem. As shown in \cite{Sandu_2015_fdvar-aposteriori,Sandu_2015_POD-inverse-problems}, the quality of the surrogate 4D-Var problem's solution depends not only on the accuracy of the surrogate model on forward dynamics, but also on the accurate modeling of the adjoint dynamics. 

As the model ($\Model$) encapsulates the known physics of the system under consideration, the model derivative ($d\Model/d\theta$) can be interpreted as known physical information. Incorporating this physical knowledge into the training of a machine learning model is known as the physics-guided or science-guided machine learning paradigm \cite{Willard2020}. Our approach incorporates this information into the neural network surrogate training by appending an adjoint mismatch term to the loss function. Training neural networks with loss functions incorporating the derivative of the network itself has been applied for the solution of partial differential equations \cite{Raissi2019} and, recently and in a more restricted setting than we propose, to the creation of surrogates to the solution of inverse problems \cite{DINO}. 

We formulate several training methods that incorporate derivative information, and demonstrate that the resulting surrogates have superior generalization performance over the traditional approach where training uses only forward model information. The new surrogates also lead to better solutions to the 4D-Var problem.

The remainder of the paper is organized as follows.  \Cref{sec:background} introduces the 4D-Var problem, solution strategies, and the use of surrogates to speed up the solution process. Relevant topics from machine learning and the neural network training problem are discussed. \Cref{sec:theory} provides a short theoretical analysis of the solution of 4D-Var with surrogate models, and show the dependence of the 4D-Var solution quality on the accuracy of the surrogate model and its adjoint. \Cref{sec:sgml-framework} introduces the science-guided machine learning framework, and formulates the science-guided approach to the 4D-Var problem. Numerical experiment setup is described in Section \ref{sec-num-experiments}. Experimental results and their discussion are presented in \Cref{sec:results-and-discussion}, and closing remarks are given in Section \ref{sec-conclusions}.

\section{Background}
\label{sec:background}
\subsection{Data Assimilation}
\label{sec:intro-da}
%
Data assimilation  \cite{asch2016data,evensen2009data,reich2015probabilistic} is the process of combining information from imperfect computer model predictions with noisy and sparse measurements of reality, to obtain improved estimates of the state (and/or parameters) of a dynamical system of interest.  
The data assimilation problem is generally solved using either statistical or variational approaches \cite{asch2016data}. The resulting improved estimate of the system state is known as the analysis. The data assimilation problem is posed in a Bayesian framework, where one starts with a known prior distribution of possible background states, and assumes a likelihood distribution of the observations conditioned by the system state. The data assimilation algorithm then aims to produce a sample, called the analysis, from the posterior distribution of system states \cite{asch2016data, SanduIntroDA}.

The data assimilation problem can be solved in either a filtering or a smoothing setting. In the filtering setting, observation information from the current time $t_{0}$ is used to improve an estimate of the system's current state at time $t_{0}$. In the smoothing setting, observation information from current and future times $t_{0}, \dots, t_{n}$ is used to produce an improved estimate of the system state at time $t_{0}$.

\begin{subequations}\label{eqn:4dvar-setting}
We now formally describe the smoothing problem. Our knowledge of the system at hand is encapsulated in the following. We have a finite dimensional representation $\danx_{i} \in \Re^\nstate$ of the state the physical system at time $t_{i}$,  $i \ge 0$, and a background (prior) estimate $\xb_{0}$ of the true system state $\xt_{0}$ at time $t_{0}$,
\begin{equation}
        \danx_{0} =\xb_{0} = \xt_{0} + \errm_{0}, \errm_{0} \from \mathcal{N}(0, \danB_{0}).
\end{equation}
We also assume that we have noisy observations, or measurements, of the state at times $t_{i}$,  $i \ge 0$,
\begin{equation}
        \dany_{i} = \Hobs_i(\xt_i) + \erro_i, \erro_{i} \from \mathcal{N}(0, \danR_{i}),
\end{equation} and a high fidelity model operator $\Model_{i-1, i}(\danx)$ which transforms the system state at time $t_{i-1}$ to one at time $t_{i}$, up to some model error $\errm_{i}$,
\begin{equation}
        \danx_{i} = \Model_{i-1,i}(\danx_{i-1}) + \errm_i,  \qquad  \errm_{i} \from \mathcal{N}(0, \danQ_{i}).
\end{equation}
We further assume access to all noise covariance matrices, namely the background noise covariance matrix $\danB_{0}$, observation noise covariance matrices $\danR_{i}$, and model noise covariance matrices $\danQ_{i}$.
\end{subequations}

\subsection{Variational solution to the smoothing problem}
A solution to the variational data assimilation problem is defined as the numerical solution to some appropriately formulated optimization problem \cite{asch2016data}. The 4D-Var approach is a solution to the smoothing problem, and an example of a general variational inverse problem. Under the perfect model assumption, the goal of the 4D-Var is to find an initial value for our dynamical system by weakly matching sparse, noisy observations along the model-imposed trajectory and prior information about the system's state  \cite{guckenheimer2013nonlinear}. 

\subsubsection{4D-Var Problem Formulation}
 4D-Var seeks the maximum a posteriori estimate $\xa_{0}$ of the initial state at $t_{0}$ subject to the constraints imposed by the high fidelity model equations:
\begin{align}\label{eqn:4dvar}
   \xa_0 &= \argmin \Jfunc(\danx_0)\nonumber \\
    \text {subject to} \quad \danx_{i} &= \Model_{i-1,i}(\danx_{i-1}), ~~i = 1, \dots, n; \\
    \Jfunc(\danx_0) &\coloneqq 
      \sfrac{1}{2}\norm{\danx_0 - \xb_{0} }_{{\danB_{0}}^{-1}}^{2} + \sfrac{1}{2} \sum_{i = 1}^{n} \norm{\Hobs_i(\danx_{i}) - \dany_{i}}_{\danR_{i}^{-1}}^{2},\nonumber
\end{align}
where $\norm{\danx}_{\mathbf{A}} \coloneqq \sqrt{\danx^{T}\mathbf{A}\danx}$, with $\mathbf{A}$ a symmetric and positive-definite matrix \cite{Sandu2019}. This corresponds to the setting described in \eqref{eqn:4dvar-setting}, with the additional assumption that there is no model error, i.e., $\danx_{i} = \Model_{i-1,i}(\danx_{i-1})$.

\subsubsection{Solving the 4D-Var Optimization Problem}
In practice, the constrained minimization problem \eqref{eqn:4dvar} is solved using gradient-based optimization algorithms. We denote the Jacobian of the model solution operator with respect to model state (called the ``tangent linear model''), and the Jacobian of the observation operator, by:
\begin{subequations}
\begin{align}
\label{eqn:TLM}
    \Mtlm_{i,i+1}(\danx_{i}) &\coloneqq \Model_{i,i+1}'(\danx_{i}) \equiv \sfrac{d \Model_{i,i+1}(\danx)}{d \danx}\Big\rvert_{\danx = \danx_{i}} \in \Re^{\Nstate \times \Nstate}, \\
\label{eqn:HTLM}
   \HH_i(\danx_{i}) &\coloneqq \Hobs_{i}'(\danx_{i}) \equiv \sfrac{d \Hobs_{i}(\danx)}{d \danx}\Big\rvert_{\danx = \danx_{i}} \in \Re^{\Nobs \times \Nstate},
\end{align}
\end{subequations}
respectively. Applying the identity $\danx_{i} = \Model_{i-1,i}(\danx_{i-1})$, $i = 1, \dots, n$, to remove the constraints in \eqref{eqn:4dvar}, the gradient of the 4D-Var cost function \eqref{eqn:4dvar} takes the form: 
\begin{eqnarray}
\label{eqn:4dvar-gradient} 
         \nabla_{\danx_0} \Jfunc(\danx_0) &=& \danB_{0}^{-1}\,(\danx_0 - \xb_{0}) + \sum_{i = 1}^n \Mtlm_{0,i}^T(\danx_{0})\, \HH_i^T\, \danR_{i}^{-1}\, (\Hobs_{i}(\danx_{i}) - \dany_{i}), \\
                 \nonumber
  \Mtlm_{0,i}^T(\danx_{0})  &=&  \prod_{k=1}^{i} \Mtlm_{i-k,i-k+1}^T(\danx_{i-k}).
\end{eqnarray}
This motivates the need for an efficient evaluation of the ``adjoint model'' $\Mtlm_{k,k+1}^T$. For our purpose in this paper, the adjoint model is the transpose of the tangent linear model, although the method of adjoints is typically derived in a more general setting \cite{asch2016data}.

The cost of solving the 4D-Var problem \eqref{eqn:4dvar} is dominated by the computational cost of evaluating the cost function $\Jfunc$ \eqref{eqn:4dvar} and its gradient \eqref{eqn:4dvar-gradient}. Evaluating $\Jfunc$ requires integrating the model $\Model$ forward in time. Evaluating $\nabla_{\danx_0} \Jfunc(\danx_0)$ requires evaluating $\Jfunc$ and running the adjoint model $\Mtlm^T$ backward in time, both runs being performed over the entire simulation interval $[t_{0}, t_{n}]$. Techniques for efficient computation of the sum in \eqref{eqn:4dvar-gradient} are based on computing only adjoint model-times-vector operations \cite{Sandu_2010_inverseDADJ, Sandu_2005_adjointAQM, Sandu_2008_fdvarTexas}. Each full gradient computation requires a single adjoint run over the simulation interval. 

\subsubsection{4D-Var with Surrogate Models}
%
Surrogate models for fast, approximate inference  have enjoyed great popularity in data assimilation research \cite{BRAJARD2020101171, popov2020multifidelity, popov2021multifidelity, Sandu_2012_quasiFdvar, Sandu_2017_SMDEIM, Sandu_2015_POD-inverse-problems}. Surrogate models are most often applied in one of two ways: to replace the high fidelity model dynamics constraints in \eqref{eqn:4dvar}, or to supplement the model by estimating the model error term $\errm_{i}$ in \eqref{eqn:4dvar-setting} \cite{Farchi2020}.  In the former approach the derivatives of the surrogate model replace the derivatives of the high fidelity model in \eqref{eqn:TLM} and in the 4D-Var gradient calculation in \eqref{eqn:4dvar-gradient}. The universal approximation property of neural networks and their relatively inexpensive forward and derivative evaluations make them a promising candidate architecture for surrogate construction \cite{aggarwal2018neural, Farchi2020, Panda2020, https://doi.org/10.48550/arxiv.2108.12344}. The work \cite{DINO} incorporates derivative information into the training of a neural network surrogate for a restricted class of inverse problems. 

This work differs from \cite{DINO} by incorporating adjoint information of the physical model directly into the surrogate model. Specifically, our surrogate model imitates the dynamical system timestepping operator and its derivative, rather than modeling the state to cost function operator. In addition, our method incorporates known derivative information to create more accurate neural network surrogate models of the physical system which can be used outside of the solution of the individual inverse problem, which can be used in any other context that cheap surrogates of the underlying physical process are required. One example would be a separate inverse problem, where, for example, the observation operator and its observation noise covariance matrix have changed, or incorporation into an ensemble-based data assimilation procedure. In these cases, our surrogates of the dynamical system could directly be incorporated into the new problem setting with no additional training, as long as the underlying dynamical system remains the same. The additional restrictions on the problem setting in \cite{DINO} potentially allow the more natural application of existing dimensionality reduction techniques for outer loop applications at the expense of fixing the neural network to the inverse problem of interest unless the network is retrained. 

\subsection{Neural Networks}
Neural networks are parametrized function approximations with architectures loosely inspired by the biological structure of brain \cite{aggarwal2018neural}. The canonical example of an artificial neural network is the feedforward network, or multilayer perceptron. Given an element-wise nonlinear activation function $\phi$, weight matrices $\{ \mathbf{W}_{i} \}_{i = 1, \dots, k}$, $\mathbf{W}_{i} \in \Re^{\Nsub{i} \times \Nsub{i-1}}$, $i = 2, \dots, k$, and bias vectors $\{\mathbf{b}_{i}\}_{i = 1, \dots, k-1}$, $\mathbf{b}_{i} \in \Re^{\Nsub{i}}$ for $i = 1, \dots, k - 1$, $\mathbf{b}_k \in \Re^{\Nsub{k}}$, the action $\zout$ of a $k$-layers feed-feedforward network on an input vector $\uin$ is given by the sequence of operations:
\begin{equation}
\label{eqn:general-NN}
\begin{split}
    \mathbf{z}_{1} &= \phi(\mathbf{W}_{1}\uin + \mathbf{b}_{1}), \\
    \mathbf{z}_{i} &= \phi(\mathbf{W}_{i}\,\mathbf{z}_{i-1} + \mathbf{b}_{i}), \quad i = 2, \dots, k - 1, \\
    \zout &= \mathbf{W}_{k}\,\mathbf{z}_{k-1} + \mathbf{b}_{k},
    \end{split}
\end{equation}
with $\mathbf{z}_{i} \in \Re^{\Nsub{i}}$ for $i = 1, \dots, k - 1$.
Popular choices of activation function include hyperbolic tangent and the rectified linear unit (ReLU) \cite{aggarwal2018neural}.  In general, the dimension of each layer's output may vary.  The final bias vector may be omitted to suggest more clearly the notion of linear regression on a learned nonlinear transformation of the data, but we retain it in our implementation. The input dimension $\uin  \in \Re^{\Nsub{\rm in}}$ and the output dimension $\uout  \in \Re^{\Nsub{\rm out}}$ are specified by the training data from the problem at hand.

\subsection{Training Neural Networks}
In the supervised learning setting, we assume access to pairs of input/output data, $\{(\uin^{\{\ell\}}, \uout^{\{\ell\}})\}_{\ell=1,\dots, \Nsub{data}}$, called the training data. Denoting the output of the neural network with weight matrices and bias vectors contained in the single parameter vector $\theta$ as $\nnm(\uin^{\{\ell\}};\theta)$, neural networks are typically trained on a generic loss function additive in the training data samples of the form
\begin{equation}\label{eqn:generic-ml-loss-function}
    \frac{1}{\Nsub{data}}\sum_{i=1}^{\Nsub{data}}\loss{}\left(\nnm(\uin^{\{i\}};\theta), \uout^{\{i\}} \right).
\end{equation}
Common examples of $\loss{}$ include $2$-norm difference between the arguments for regression problems and categorical cross-entropy for categorization. The structure of neural networks and additive loss function allows use of the backpropagation algorithm for efficient gradient computation of losses of the form \Cref{eqn:generic-ml-loss-function} with respect to the training data \cite{aggarwal2018neural}. This allows easy use of the standard algorithm for training neural networks, stochastic gradient descent, where an estimate of the loss function across the entire dataset and its gradient with respect to the network weights is computed using only a sample of training data, i.e with the summation in \Cref{eqn:generic-ml-loss-function} is instead over a subset of size $\Nsub{batch}$ of $\{1, 2, \dots, \Nsub{data}\}$, typically chosen uniformly without replacement \cite{aggarwal2018neural}. Stochastic gradient descent is often combined with an acceleration method such as Adam, which makes use of estimates of the stochastic gradient's first and second order moments to accelerate convergence \cite{Kingma2014}. Constraints can be enforced in the neural network training process through adding a penalty term to the loss function, or through modifying a traditional constrained optimization algorithm to work in the stochastic gradient descent setting \cite{Dener2020}.

\subsection{Neural Networks in Numerical Weather Prediction}
Different applications of neural network models in the context of numerical weather prediction have been explored. Previously studied approaches include learning of the model operator by feedforward networks \cite{gmd-11-3999-2018}, learning repeated application of the model operator by recurrent neural network for use in 4D-Var \cite{Panda2020}, approaches based on learning the results of the data assimilation process \cite{app11031114}, and online and offline approaches for model error correction \cite{farchi2021comparison}. While to the authors' knowledge, deep-learning based models have failed to match operational weather-prediction models in prediction skill \cite{gmd-11-3999-2018, farchi2021comparison, canmachineslearn} they nonetheless have offered comparatively inexpensive approximations of operational operators which can be applied to specific areas of interest. They have thus been identified as an area of application for machine learning by organizations such as the European Centre for Medium-Range Weather Forecasts \cite{ecmwfAI}.

\section{Theoretical Motivation}
\label{sec:theory}
It has been shown in \cite{Sandu_2015_fdvar-aposteriori, Sandu_2012_quasiFdvar, Sandu_2015_POD-inverse-problems} that when the optimization \eqref{eqn:4dvar} is performed with a surrogate model $\nnm$ in place of the model operator $\Model$, the accuracy of the resulting solution depends both on accuracy of the surrogate forward model and the accuracy of its adjoint. A rigorous derivation of error estimates in the resulting 4D-Var solution upon the forward and adjoint surrogate solutions has developed in \cite{Sandu_2015_fdvar-aposteriori}. We will follow a simplified analysis in the same setting to  demonstrate the requirement for accurate adjoint model dynamics.

The constrained optimization problem \eqref{eqn:4dvar} can be solved analytically by the method of Lagrange multipliers. The Lagrangian for \eqref{eqn:4dvar} is:
\begin{equation}
\begin{split}\label{eqn:lagrangian}
    \mathtt{L}(\danx_{0:n}; \llam_{0:n}) &= \sfrac{1}{2}\norm{\xb_{0} - \danx_{0}}_{\danB_{0}^{-1}}^{2} + \sfrac{1}{2} \sum_{i = 1}^{n} \norm{\Hobs_i(\danx_{i}) - \dany_{i}}_{\danR_{i}^{-1}}^{2} \\ &+ \sum_{i = 0}^{n - 1}\llam_{i+1}^{T}\big(\danx_{i + 1} - \Model_{i, i + 1}(\danx_{i})\big),
\end{split}
\end{equation}
where $\danx_{0:n}$ is the vector containing all states, and $\llam_{0:n}$ is the vector containing all Lagrange multipliers, with $\llam_{i+1}$ associated with the constraint $\danx_{i + 1} - \Model_{i, i + 1}(\danx_{i}) = 0$. A local optimum to the constrained optimization problem \eqref{eqn:4dvar} is stationary point of \eqref{eqn:lagrangian} and therefore fulfills the first order optimality necessary conditions:
\begin{subequations}
\label{eqn:full-adjoint}
\begin{align}
    \begin{split}
    &\text{\it High fidelity forward model:} \\ 
    &\quad \danx_0 = \xa_0;  \\
    &\quad\danx_{i+1} = \Model_{i, i+1}(\danx_{i}), \quad i = 0, \dots, n - 1;
    \end{split} \\
    \begin{split}
    &\text{\it High fidelity adjoint model: } \\
    &\quad\llam_{n} = \HH_{n}^{T}\,\danR_{n}^{-1}\,(\dany_{n} - \Hobs_{n}(\danx_{n})); \\
    &\quad\llam_{i} = \Madjt_{i+1, i}\llam_{i+1} + \HH_{i}^{T}\danR_{i}^{-1}(\dany_{i} - \Hobs_{i}(\danx_{i})), \\ 
    &\quad i = n-1, \dots, 0;
    \end{split} \\
    \begin{split}
    &\label{eqn:fullKKTgrad}\text{\it High fidelity gradient: } \\ 
    &\quad\nabla_{\danx_{0}} \Psi(\xa_{0}) = \danB_{0}^{-1}\,(\xa_{0} - \xb_{0}) - \llam_{0} = 0.
    \end{split}
\end{align}
\end{subequations}
Suppose we have some differentiable approximation $\nnm$ to our high fidelity model operator $\Model$. We have:
\begin{subequations}
\label{eqn:surrogate-model}
\begin{equation}
\label{eqn:surrogate-model-fwd}
\nnm_{i, i +1}(\danx) = \Model_{i, i+1}(\danx) + \dane_{i, i+1}(\danx), 
\end{equation}
where the surrogate model forward approximation error is $\dane_{i, i+1}(\danx) = \nnm_{i, i+1}(\danx) - \Model_{i, i+1}(\danx)$. Following \eqref{eqn:TLM} the tangent linear surrogate model is:
\begin{equation}
\label{eqn:surrogate-model-tlm}
\quad \mathbf{N}_{i, i +1}(\danx) \coloneqq \nnm_{i, i +1}'(\danx) = \Mtlm_{i, i+1}(\danx) + \dane_{i, i+1}'(\danx). 
\end{equation}
\end{subequations}
Replacing $\Model$ by $\nnm = \Model + \dane$ in \eqref{eqn:4dvar} leads to the perturbed 4D-Var problem:
\begin{equation}
\begin{split}
\label{eqn:4dvar-pert}
        	\danx^{a*}_0 &= \argmin \Jfunc(\danx_0)\\ 
        	\text {subject to} \quad \danx_{i} &= \nnm_{i,i+1}(\danx_{i}), \quad
        	i = 0, \dots, n-1,
\end{split}
\end{equation}
and its corresponding optimality conditions:
\begin{subequations}
\label{eqn:reducedKKT}
\begin{align}
    \begin{split}
    &\text{\it Surrogate forward model:} \\
    &\quad\danx_{0}^{*} = \danx_{0}^{a*}; \\
    &\quad\danx^{*}_{i+1} = \nnm_{i, i+1}(\danx_{i}^{*}), \quad i = 0, \dots, n - 1;
    \end{split} \\ 
    \begin{split}
    &\text{\it Surrogate adjoint model: } \\
    &\quad\llam^{*}_{n} = \HH_{n}^{T}\danR_{n}^{-1}(\dany_{n} - \Hobs_{i}(\danx_{n}^{*})); \\
    &\quad\llam^{*}_{i} = \mathbf{N}_{i, i+1}^{T}\llam^{*}_{i+1} +\HH_{i}^{T}\danR_{i}^{-1}(\dany_{i} - \Hobs_{i}(\danx_{i}^{*})),\quad
     i = n-1, \dots, 0;
    \end{split}\\
    \begin{split}
    &\text{\it Surrogate gradient: }\\
    &\quad\nabla_{\danx_{0}} \Psi(\danx_{0}^{a*}) = \danB_{0}^{-1}(\danx_{0}^{a*} - \danx_{0}^{b}) - \llam^{*}_{0} = 0.
    \end{split}
\end{align}
\end{subequations}

Let $\danx_{0}^{a*}$ be a solution to the surrogate 4D-Var problem \eqref{eqn:4dvar-pert} satisfying its corresponding optimality conditions and $\xa$ a solution to the full 4D-Var problem \eqref{eqn:4dvar}.  For the sake of analysis, we assume that the solution to the 4D-Var and perturbed 4D-Var problems are unique. We determine the quality of $\danx^{a*}_{0}$ as a solution to the original 4D-Var problem \eqref{eqn:4dvar} by examining the difference of the two solutions' first-order optimality conditions. Using \eqref{eqn:surrogate-model} in \eqref{eqn:reducedKKT} and subtracting \eqref{eqn:full-adjoint} leads to:
\begin{subequations}
\begin{align}
    \begin{split}
    &\text{\it Forward model error:} \\
    &\quad \danx_{0}^{*} - \danx_{0}  = \Delta\xa_{0}~~(\text{see}~\eqref{eqn:KKTsoldiff}), \\
    &\quad \danx_{i}^{*} - \danx_{i} = \Model_{i-1, i}(\danx_{i-1}^{*}) - \Model_{i-1, i}(\danx_{i-1}) + \dane_{i-1, i}(\danx_{i-1}^{*}), \\
    &\quad i = 1, \dots, n. \end{split}\\ 
    \begin{split}
    \label{eqn:KKTlagrangemultdiff}
    &\text{\it Adjoint model error: } \\
    &\quad \llam_{n}^{*} - \llam_{n} = \HH_{n}^{T}\danR_{n}^{-1}(\H_{n}(\danx_{n}) - \H_{n}(\danx_{n}^{*})), \\
    &\quad \llam_{i}^{*} - \llam_{i} = \Madjt_{i+1, i}(\llam_{i+1}^{*} - \llam_{i+1}) + \HH_{i}^{T}\danR_{i}^{-1}(\H_{i}(\danx_{i}) - \H_{i}(\danx_{i}^{*})) \\
    &\qquad\qquad + \dane_{i, i+1}'(\danx_{i}^{*})^{T}\llam^{*}_{i + 1}, \quad i = n-1, \dots, 0. 
    \end{split}\\
    \begin{split}
    \label{eqn:KKTsoldiff}
    &\text{\it Optimal solution error: } \\
    &\quad \Delta\xa_{0} \equiv \danx_{0}^{a*} - \xa_{0} = \danB_{0}(\llam_{0}^{*} - \llam_{0}).
    \end{split}
\end{align}
\end{subequations}

We see that the additional error at each step of the forward evaluation depends on the difference of the model operator applied to $\danx_{i}$, the accumulated error, and the mismatch function $\dane$ itself. The error in the adjoint variables depends on the forward error as transformed by the observation operator and the adjoint of the mismatch function applied to specific vectors, namely the perturbed adjoint variables $\llam_{i}^{*}$. Since the error of the solution the mismatch in final adjoint variables, we see that the quality of the solution depends directly on both the mismatch function $\dane$ and its Jacobian $\dane'$. 

We now discuss which adjoint-vector products supply the most useful information to the neural network surrogate for the solution of the 4D-Var problem. Equation \eqref{eqn:KKTsoldiff} indicates that the accuracy of the optimal solution depends on the accumulated error in the Lagrange multiplier terms $\llam_{0}^{*} - \llam_{0}$. From \eqref{eqn:KKTlagrangemultdiff}, we see that the error in Lagrange multiplier terms accumulated from timestep $i$ to $i + 1$ depends on three terms: (a) $\Madjt_{i+1, i}(\llam_{i+1}^{*} - \llam_{i+1})$, the Lagrange multiplier error accumulated up to step $i + 1$, (b) $\HH_{i}^{T}\danR_{i}^{-1}(\H_{i}(\danx_{i}) - \H_{i}(\danx_{i}^{*}))$, which depends only on the accumulated forward model error, and (c) $\dane_{i, i+1}'(\danx_{i}^{*})^{T}\llam^{*}_{i + 1}$, the error in the surrogate adjoint model applied to the perturbed 4D-Var Lagrange multiplier term. This suggests two ways to minimize the error in the $i$-th Lagrange multiplier term $\llam_{i+1}^{*} - \llam_{i+1}$: a more accurate forward model, which reduces the magnitude of term (b), and a more accurate adjoint model, specifically the adjoint-vector product $(\Mtlm_{i, i+1} + \dane_{i, i+1}'(\danx_{i}^{*}))^{T}\llam^{*}_{i+1}$, which will reduce the magnitude of term (c). This in turn implies a more accurate solution to the optimization problem in our analytical framework. Unfortunately, these vectors are only implicitly defined in terms of the solution of the training process itself. Therefore, a computationally tractable approximation must be chosen. Several approximations are evaluated in \Cref{sec:adjv-methods}.

\section{Loss Function Construction and Science-Guided ML Framework}
\label{sec:sgml-framework}

\subsection{Traditional Supervised Learning Approach}\label{sec:traditional-network}
Neural networks offer one approach to building computationally inexpensive surrogate models for the model solution operator $\Model$ in eq. \eqref{eqn:4dvar}, which can be used in place of the high fidelity model in the inversion procedure \cite{Panda2020, Willard2020}. A neural network surrogate model \eqref{eqn:surrogate-model} takes as input the system state $\danx_{i}$ and outputs an approximation of the state advanced by a traditional time stepping method:
\begin{equation}
\label{eqn:NN-surrogate-model}
\nnm(\danx_{i};\theta) = \Model_{i, i+1}(\danx) + \dane_{i, i+1}(\danx), 
\end{equation}
where $\theta$ denote the vector of parameters of the neural network. In our methodology, the neural network surrogate \eqref{eqn:NN-surrogate-model} is fixed and does not vary between time steps, therefore we do not use time subscripts. Moreover, the same set of parameters $\theta$ is used for all time steps $i$. Because our experiments involve training a neural network to emulate the solution of an autonomous system of ordinary differential equations, here we do not include time as a model input or switch between multiple surrogate models.

\begin{figure*}[h]
\centering
\includegraphics[width=0.8\textwidth]{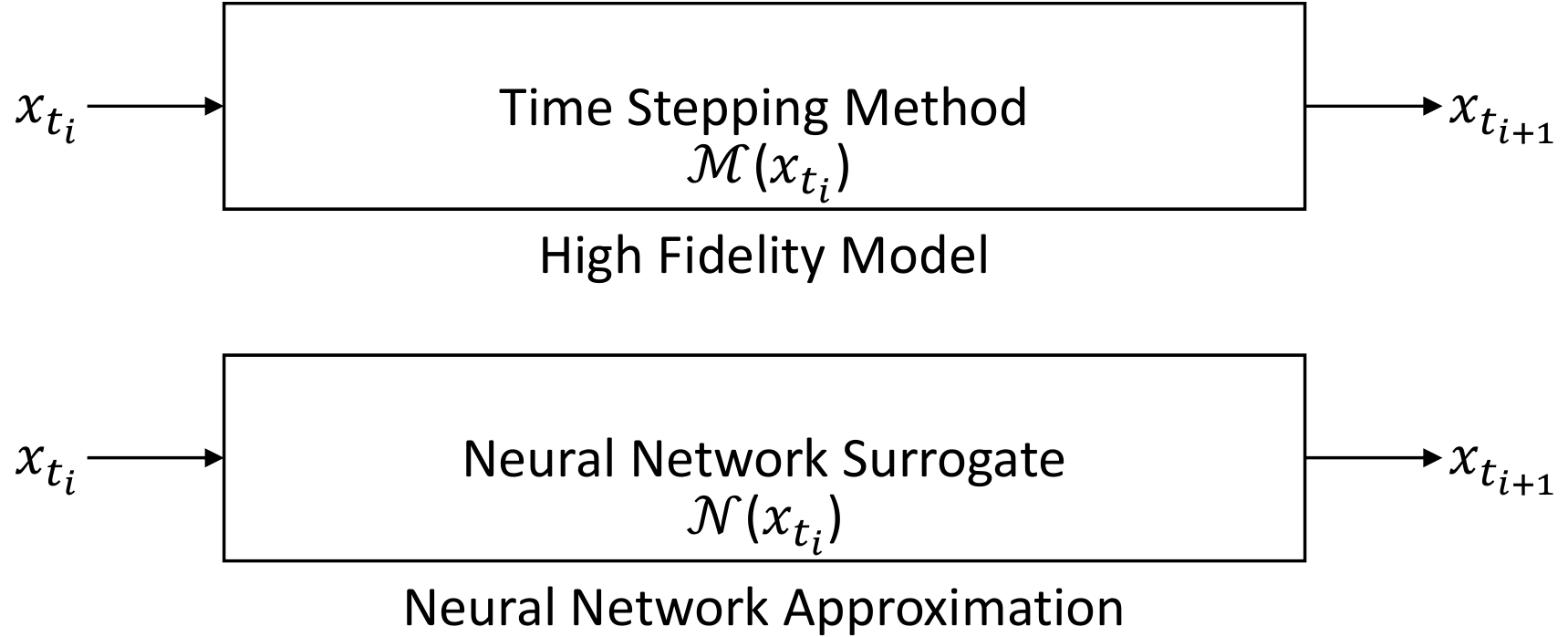}
\caption{Diagram of the neural network surrogate approach. A time stepping method advances the system state according to a scientifically derived formula. The neural network learns to provide a cheap surrogate of the high fidelity timestepping method through the training process.} 
\label{surrogate-image}
\end{figure*}

The standard approach is to train the surrogate model $\mathcal{N}$ on input-output data pairs 
\begin{equation}\label{eqn:data}
       \left\{\big(\danx_{t_{i}},\danx_{t_{i+1}} = \Model_{t_{i}, t_{i+1}}(\danx_{t_{i}})\big) \right\}_{i = 1, \cdots, \Nsub{data}}
\end{equation}
collected from trajectories of the high fidelity model \eqref{eqn:4dvar-setting}. We have replaced the subscript $i$ used earlier with a double subscript $t_{i}$ to indicate that the data used in the training process need not consist of all snapshots collected from a single model trajectory. Training is accomplished by minimizing squared two-norm mismatch summed across the  training data set:
\begin{equation}\label{eqn:loss-standard}
        \loss{Standard} (\theta) \coloneqq \frac{1}{\Nsub{data}}\sum_{i=1}^{\Nsub{data}} \norm{\mathcal{N}(\danx_{t_{i}};\theta) - \Model_{t_{i}, t_{i+1}}(\danx_{t_{i}})}_2^2.
\end{equation}
%

\subsection{Adjoint-Match Training}
Incorporating known physical quantities into  the training of a neural network by an additional term in the loss function is one of the basic methods of increasing model performance on complex scientific data \cite{Willard2020}. In the variational data assimilation context, we assume access to the high fidelity model operator's adjoint. Our goal is to devise a training method that incorporates this adjoint information and produces more accurate solutions to the surrogate 4D-Var problem. 
We denote the Jacobian \eqref{eqn:surrogate-model-tlm} of the neural network surrogate by
\begin{equation}
\label{eqn:nnTLM}
    \nnmtlm(\danx_{i};\theta) \coloneqq \sfrac{\partial \nnm(\danx;\theta)}{\partial \danx}\Big\rvert_{\danx = \danx_{i}} \in \Re^{\Nstate \times \Nstate}.
\end{equation}
A natural means of incorporating adjoint information into the training process is to add the adjoint operator mismatch to the cost function
\begin{equation}
\label{eqn:loss-adjoint}
\begin{split}
    \loss{Adj}(\theta) &\coloneqq \loss{Standard} (\theta) + \sfrac{\alpha}{\Nsub{data}}\, \sum_{i=1}^{\Nsub{data}}  \norm{ \nnmtlm^{T}(\danx_{t_{i}};\theta) - \Madjt_{t_{i}, t_{i+1}}(\danx_{t_{i}})}_F^2 \\
    &= \loss{Standard} (\theta) + \sfrac{\alpha}{\Nsub{data}}\, \sum_{i=1}^{\Nsub{data}}  \norm{ \nnmtlm(\danx_{t_{i}};\theta) - \Mtlm_{t_{i}, t_{i+1}}(\danx_{t_{i}})}_F^2,
    \end{split}
\end{equation}
where the regularization parameter $\alpha$ determines how heavily the adjoint mismatch term is weighted. We note that the term $\norm{ \nnmtlm^{T}(\danx_{t_{i}};\theta) - \Madjt_{t_{i}, t_{i+1}}(\danx_{t_{i}})}_F^2$ in \eqref{eqn:loss-adjoint} is exactly equal to $\norm{ \nnmtlm(\danx_{t_{i}};\theta) - \Mtlm_{t_{i}, t_{i+1}}(\danx_{t_{i}})}_F^2$, but the computation of the model adjoint operator $\Madjt_{t_{i}, t_{i+1}}(\danx_{t_{i}})$ requires considerably more computational complexity than computation of the model tangent linear model  $\Mtlm_{t_{i}, t_{i+1}}(\danx_{t_{i}})$ due in part to the additional memory requirements \cite{SanduAdjoints}. Later, we will explore training with adjoint-vector products rather than the full adjoint matrix itself. Accurate adjoint-vector products at the local optima, along with accuracy of the forward model, is the necessary condition for accurate solution to the surrogate-optimized problem derived in \cite{Sandu_2015_fdvar-aposteriori} and in \Cref{sec:theory}. We therefore use the formulation \eqref{eqn:loss-adjoint}. Although we do not consider the time of the training process in this paper, the evaluation of \eqref{eqn:loss-adjoint} requires evaluation of the neural network's adjoint, and  thus increases the computational expense of the training problem. The cost of evaluating the network and its derivative after the training process is not affected by the choice of training cost function.

Training neural networks using loss functions  that incorporate the derivative of the network itself was first attempted in \cite{sobolevTraining}, has been applied for the solution of partial differential equations \cite{Raissi2019}, and, more recently, to the solution of inverse problems \cite{DINO}.

\subsection{Training with Adjoint-Vector Products}\label{subsec-adjv-training}
In operational settings, adjoint-vector products $\Madjt \mathbf{v}$ are more readily available than the full high fidelity model adjoint operator $\Madjt$. Using our example from \Cref{sec:intro}, for $10^9$ variables one dense Jacobian matrix of the full model operator represented in double precision floating point format would require $10^{9}\cdot 10^{9} \cdot 64$ bits, or $8$ exabytes of memory to store. Instead of training our network with full adjoint matrix mismatch \eqref{eqn:loss-adjoint}, we consider training the surrogate to match adjoint-vector products of the physical model on a given set of vectors $\{\mathbf{v}_{i}\}_{i=1}^{\Nsub{data}}$, which results in the following loss function:
\begin{align}\label{eqn:loss-adjvec}
    \begin{split}\loss{AdjVec}(\theta) &\coloneqq \loss{Standard} (\theta) + \sfrac{\alpha}{\Nsub{data}}\, \sum_{i=1}^{\Nsub{data}} \norm{ \nnmtlm(\danx_{t_{i}};\theta)^{T}\mathbf{v}_{i} - \Madjt_{t_{i}, t_{i+1}}\mathbf{v}_{i}}_2^2.\end{split}
    \end{align}

We note that like the adjoint training procedure \eqref{eqn:loss-adjoint}, evaluation of \eqref{eqn:loss-adjvec} requires evaluation of neural network adjoint-vector products, and thus increases computational expense of the training procedure over using the standard loss function \eqref{eqn:loss-standard}. The post-training costs of neural network and neural network adjoint evaluations are not affected.

\subsubsection{Choice of Vectors for Adjoint-Vector Products}
\label{sec:vector-choice}
To construct the adjoint-vector loss \eqref{eqn:loss-adjvec} one builds terms of the form
$(\nnmtlm^{T} - \Madjt_{t_{i}, t_{i+1}})\,\mathbf{v}_{i}$. We now discuss the selection of the directions $\mathbf{v}_{i}$ along which the two adjoints are similar. The theoretical analysis in \Cref{sec:theory} is used to inform which adjoint-vector products are expected to have the greatest impact on increasing accuracy of the solution to the surrogate-model 4D-Var problem. 

Accurate solution to the 4D-Var problem  requires small errors $\dane_{i, i+1}$ in forward model surrogate, and small errors $\mathbf{N}_{i, i+1}^{T} \llam^{*}_{i+1}- \Mtlm_{i, i+1}^{T}\llam_{i+1}$ in the surrogate adjoint-vector products. During the training process, $\llam^{*}_{i+1}$ is defined only implicitly, since it is derived from the surrogate 4D-Var problem using the trained model. For feasibility of obtaining training data, we make the approximation $\llam^{*}_{i+1}\approx\llam_{i+1}$, where $\llam_{i+1}$ is the Lagrange multiplier term from the original, high-resolution 4D-Var Problem.  A natural choice of the vectors are the adjoint solutions $\mathbf{v}_{i} = \llam_{t_i+1}$ \eqref{eqn:adjoint-vector-form}. This can be easily computed offline during the training data generation phase \eqref{eqn:data} with only one adjoint model run per calculation. A randomized approximation is obtained as follows. The computation of the 4D-Var gradient \eqref{eqn:4dvar-gradient} requires the evaluation of terms of the form
\begin{equation}
\label{eqn:adjoint-vector-form}
    \llam_m = \sum_{i = 1}^m 
     \Mtlm_{0,i}^T  \, \HH_i^T\, \danR_{i}^{-1}\, \big(\Hobs(\danx_{i}) - \dany_{i}\big), \quad m = n,\dots,0,
\end{equation}
where the adjoint model multiplies scaled innovation vectors. For operational systems the model operator and the observation noise covariance matrices $\danR_{i}$ are known.
Using the relation between the state, observations, and observation error covariance in the 4D-Var problem setting \eqref{eqn:4dvar-setting}, we can generate an artificial observation $\hat{\dany}_{t_{i}} \coloneqq \Hobs(\danx_{t_{i}}) + \erro_{t_{i}}$ with $\erro_{t_{i}} \from \mathcal{N}(0, \danR_{t_{i}})$ to substitute for $\dany_{t_{i}}$ in \eqref{eqn:adjoint-vector-form}. This leads to the following computable choice of vectors:
\begin{equation}\label{eqn:adjoint-vector-computable-form}
    \mathbf{v}_{i} = \HH_{t_{i} + 1}^T\, \danR_{t_{i} + 1}^{-1}\, \big(\Hobs(\danx_{t_{i} + 1}) - \hat{\dany}_{t_{i} + 1}\big) = \HH_{t_{i} + 1}^T\, \danR_{t_{i} + 1}^{-1}\,  \erro_{t_{i} + 1}.
\end{equation}
We note that the observation operator and observation noise covariance matrix are assumed to be available for the solution to the 4D-Var problem, so their use in the neural network training process only represents the incorporation of readily-accessible information in operational settings. We refer to this method as {\tt Lagrange} in \Cref{sec:adjv-methods}.

A second approach is to consider random vectors $\mathbf{v}_{i} \from \mathcal{N}(0, \Id)$  sampled from a standard normal distribution that does not take into account application-specific data. The {\tt Random} method is trained with these vectors in \Cref{sec:adjv-methods}. Finally, a third approach is to use random standard basis vectors. These vectors are used to train the {\tt RandCol} method in \Cref{sec:adjv-methods}.

\subsection{Independent Forward/Adjoint Surrogate Training}
Another machine learning-based approach to model reduction is the construction of separate surrogate models of the forward and for the adjoint dynamics. Use of a regularized cost function as in our original formulation \eqref{eqn:loss-adjoint} results in an inherent trade-off between minimizing the forward data mismatch term and the weighted penalty term. In principle, separate models may allow a more accurate modeling of the forward and adjoint dynamics. We construct two neural networks. The first, $\nnm_{\rm IndepFwd}(\danx;\theta_{\rm Fwd})$, models only the forward dynamics and is trained with the loss function:
\begin{equation}\label{LossIndepFwd}
        \loss{IndepFwd}(\theta) \coloneqq \frac{1}{\Nsub{data}}\sum_{i=1}^{\Nsub{data}} \norm{\nnm_{\rm IndepFwd}(\danx_{t_{i}};\theta) - \Model_{t_{i}, t_{i+1}}(\danx_{t_{i}})}_2^2.
\end{equation}
The second one, $\nnm_{\rm IndepAdj}(\danx;\theta_{\rm Adj})$, models the adjoint dynamics of the high resolution model. Although the adjoint $\Madjt_{t_i}(\danx_{t_{i}})$ is itself a linear operator, the function $\danx_{t_{i}} \mapsto \Madjt_{t_i}(\danx_{t_{i}})$ in general is not, justifying use of a nonlinear approximation. The adjoint network is trained with the loss function:
\begin{equation}\label{LossIndepAdj}
        \loss{IndepAdj}(\theta) \coloneqq \frac{1}{\Nsub{data}}\sum_{i=1}^{\Nsub{data}} \norm{\nnm_{\rm IndepAdj}(\danx_{t_{i}};\theta) - \Madjt_{t_{i}, t_{i+1}}(\danx_{t_{i}})}_F^2.
\end{equation}
In \Cref{LossIndepAdj}, we have slightly abused notation by taking the difference of the network's output, a vector in $\Re^{\Nstate \cdot \Nstate}$, and the adjont matrix $\Madjt_{t_{i}, t_{i+1}}(\danx_{t_{i}})$, a matrix in $\Re^{\Nstate \times \Nstate}$. A reshape operation must be applied when the network is implemented.

The input to the trained network $\nnm_{\rm IndepAdj}$ is a state vector $\danx_{t_{i}}$ and its output a $\Nstate \times \Nstate$ matrix which approximates $\Madjt_{t_{i}, t_{i+1}}(\danx_{t_{i}})$. For the same reasons explained in \Cref{subsec-adjv-training}, training with full adjoint matrices may not be feasible in some operational settings. The independent adjoint network may also be calculated by training with the loss function formulated in terms of adjoint-vector products:
\begin{equation}\label{LossIndepAdjVec}
        \loss{IndepAdjVec}(\theta) \coloneqq \\
        \frac{1}{\Nsub{data}}\sum_{i=1}^{\Nsub{data}} \norm{\nnm_{\rm IndepAdj}(\danx_{t_{i}};\theta)\mathbf{v}_{i} - \Madjt_{t_{i}, t_{i+1}}(\danx_{t_{i}})\mathbf{v}_{i}}_2^2,
\end{equation}
where we have again slightly abused notation by using $\nnm_{\rm IndepAdj}(\danx_{t_{i}};\theta)$ to denote the $\Nstate \times \Nstate$ matrix formed by reshaping the network's $\Nstate \cdot \Nstate$ vector output.

\section{Numerical Experiments}
\label{sec-num-experiments}

\subsection{Lorenz-63 System}

A standard test problem for data assimilation is the Lorenz '63 system \cite{lorenz1963deterministic}, which was introduced by Edward N. Lorenz. It is a three variable dynamical system representing atmospheric convection, that exhibits deterministic chaos for certain choices of model parameters. Its equations are given by
\begin{equation}
\label{eqn:Lorenz63}
\begin{split}
    x' &= \sigma\,(y-x), \\
    y' &= x\,(\rho - z)-y, \\
    z' &= x\,y-\beta\, z, 
\end{split}
\qquad \textnormal{with } \sigma = 10,~ \rho = 28,~ \beta = 8/3,
\end{equation}
with the canonical values of $\sigma, \rho, \beta$ that give rise to chaotic dynamics. Our experiments use the Lorenz '63 \eqref{eqn:Lorenz63} implementation from the ODE Test Problems software package \cite{otpsoft, roberts2021ode}.

\subsection{4D-Var Problem}\label{sec:4dvar-experiment}
\subsubsection{Ground Truth Trajectory Generation}\label{sec:ground-truth-generation}
The ground truth trajectory is generated as follows. We begin with the fixed initial condition 
\begin{equation}\label{eqn:fixed-l63-initial-condition}
 \mathbf{x}_0  =  \begin{bmatrix}
  -10.0375 \\
   -4.3845 \\
   34.6514
   \end{bmatrix}.
\end{equation}
The ground truth trajectory is generated by propagating this initial condition forward $550$ timesteps of length $\Delta T = 0.12$ (units) using the explicit fourth-order Runge-Kutta method with step size $\Delta t = \Delta T/50$ (units), i.e. $50$ steps per interval, to evolve the system described in \Cref{eqn:Lorenz63}.

\subsubsection{Observations}
The 4D-Var formulation uses noisy observations collected at intervals of $\Delta T = 0.12$ model time units. We observe the first and third system states, i.e., $\danx = [x,y,z]^T$ and $\Hobs_{i}(\danx) = [x,z]^T$, with the observation operator fixed for all timesteps. The observation noise covariance matrix $\danR_{i} = \mathbf{I}_{2}$, the $2\times 2$ identity matrix, is also fixed. This covariance matrix is used both in generating the observations and in the inversion procedure. The 4D-Var window is set to $n = 2$, meaning, the 4D-Var cost function \eqref{eqn:4dvar} uses observations at two time points $t_1 = t_0 + \Delta T$ and  $t_2 = t_0 + 2\Delta T$ ahead of the initial condition. 

\subsubsection{4D-Var Model Operator}\label{sec:4dvar-model-operator}
The time high-fidelity time integration within the 4D-Var cost function is accomplished with the explicit fourth order Runge-Kutta method with step size $\Delta t = \Delta T/50$. This matches the time integration routine and settings used in \Cref{sec:ground-truth-generation}.  Discrete Runge-Kutta adjoints from \cite{SanduAdjoints} are used to compute the adjoint of the time integration. The 4D-Var method using this model operator is denoted {\tt Exact} in \Cref{sec:NN-surrogate-methods}. 

\subsubsection{Background Covariance}
The background covariance matrix $\danB_{0}$ in \eqref{eqn:4dvar-setting} is chosen to be a scaled estimate of the climatological covariance 
\begin{equation}\label{covmatrix} \danB_{0} = 
    \begin{bmatrix}
      12.4294  & 12.4323  & -0.2139 \\
   12.4323  & 16.0837  & -0.0499 \\
   -0.2139 &  -0.0499  & 14.7634
    \end{bmatrix}.
\end{equation}
The covariance matrix $\danB_{0}$ was obtained by Monte Carlo experiments on the Lorenz '63 system with the same time integration settings described above. The experimental covariance was then scaled to provide optimal RMSE for the sequential 4D-Var problem described in this section.

\subsubsection{Sequential 4D-Var Setup}\label{sec:sequential-4dvar-setup}
Our primary experiment is a sequential data assimilation problem. In the sequential data assimilation setting, the analysis at time $t_{i}$ computed by solving \Cref{eqn:4dvar} is propagated forward using the high-fidelity model operator to produce the background state for the next step  $t_{i+1} = t_{i} + \Delta T$ for $i=0,\dots,\Nsub{time}-1$. The resulting 4D-Var problem is then solved to produce a new analysis at time $t_{i+1}$. This process is repeated until the number of assimilation steps desired is reached. Our sequential data assimilation run has a total length of run of $\Nsub{time} = 550$, with an assimilation step at each $t_{i}$. The resulting analysis trajectory is compared to the ground truth trajectory described in \Cref{sec:ground-truth-generation}. These runs of $550$ steps of length $\Delta T$ are used for all 4D-Var solution accuracy tests, with the analyses from the first $50$ steps discarded in error calculations to remove noise in the solution accuracy resulting from the random initial conditions. The initial background state $\xb$ at the first iteration of the sequential 4D-Var run is calculated by perturbing the ground truth system state, found in \Cref{eqn:fixed-l63-initial-condition}, by noise randomly sampled from $\mathcal{N}(0, \danB_{0})$.

The authors note for clarity that although solution of the surrogate 4D-Var problem is an optimization problem involving neural network surrogates, these optimizations are entirely separate from the network training process. The network weights are trained offline and held constant through the sequential data assimilation test. 

\subsubsection{4D-Var Optimization Routine}
The 4D-Var data assimilation experiments are repeated in the same setting with the high fidelity model and with each of the trained machine learning models. The corresponding optimization problems \eqref{eqn:4dvar}  are solved using BFGS as implemented in MATLAB's {\it fminunc} routine, with default settings and user-specified gradients. In \Cref{sec:NN-surrogate-methods}, {\tt Forecast} denotes the forward propagation of the initial randomized background state estimate with no data assimilation performed. No optimizations are performed for this method.  

\subsubsection{4D-Var Solution Methods}
 {\tt Exact} refers to analyses computed by solving the 4D-Var problem \eqref{eqn:4dvar} using the time integration and discrete adjoints described in \Cref{sec:4dvar-model-operator}. This represents the maximum baseline accuracy the surrogate methods can reasonably hope to achieve. {\tt Adj}, {\tt AdjVec},  {\tt Indep}, {\tt IndepVec}, and {\tt Standard} are 4D-Var solutions computed using surrogate 4D-Var and gradient equations where where the high fidelity model operator $\Model$ in \eqref{eqn:4dvar} has been replaced by $\nnm$, as in \Cref{eqn:4dvar-pert}. In {\tt Adj}, {\tt AdjVec}, and {\tt Standard}, the high-fidelity adjoint $\Madjt$ is replaced by the neural network surrogate's model adjoint $\nnmadjt$ when the 4D-Var gradient is computed. In {\tt Indep} and {\tt IndepVec},  adjoint networks independent compute an estimate of the high fidelity adjoint $\Madjt$. 
The analysis from step $t_{i - 1}$ is propagated forward using the high-fidelity model operator, and is then used as the background state at time $t_{i}$. The different solution methods are summarized in \Cref{table-surrogate-methods}.

\subsubsection{Measurement of 4D-Var Analysis Solution Quality}
Accuracy of the 4D-Var solution is our primary metric metric to quantify the quality of the trained model, since it is constructed specifically to provide fast solutions to the 4D-Var problem. More specifically, we compute the spatiotemporal root mean square error (RMSE) between the 4D-Var analyses using different surrogates and the reference trajectory. Given the 4D-Var analysis trajectory $\xa_{1:\Nsub{time}}$, and the reference trajectory $\xt_{1:\Nsub{time}}$, the spatiotemporal RMSE is
\begin{equation}\label{eqn:RMSE}
\text{RMSE} = \sqrt{\sum_{i=1}^{\Nsub{time}} \frac{\norm{\xa_{i} - \xt_{i}}_{2}^{2}}{\Nsub{time}\cdot\Nstate}}.
\end{equation}
As outlined in \Cref{sec:sequential-4dvar-setup}, the summation in \Cref{eqn:RMSE} for our 4D-Var test is from $i = 50$ to $\Nsub{time} = 550$.
\subsection{Neural Networks}
\subsubsection{Neural Network Architectures}
\begin{table}
    \centering
    \caption{Summary of neural network surrogate model construction methods and baseline methods used for comparison. These methods are explored in \Cref{sec:results-and-discussion}.}
    \label{table-surrogate-methods}
    \begin{tabularx}{\textwidth}{| l | X | l | l |}
    \hline
   Method & Description & Loss Function & $\alpha$ \\ \hline
   \mycell{{\tt Forecast}\\ (No network)} & No assimilation is performed. The initial background guess at the test start is propagated forward with the high-fidelity time-stepping routine and no corrections. & None & None \\  \hline
   
   \mycell{{\tt Exact} \\ (No network)} & Assimilation is performed using the high-fidelity time-stepping routine and its exact adjoint. & None & None \\  \hline
   
   \mycell{{\tt Adj} \\ ($1$ Network)} & Forward model operator data points and corresponding full adjoint matrices used to train one network. & \eqref{eqn:loss-adjoint} & $100/3$ \\  \hline
   
   \mycell{{\tt AdjVec} \\ ($1$ Network)} & Forward model operator data with corresponding adjoint data provided as matrix-vector products, with the vectors sampled randomly from the standard Gaussian distribution, used to train one network. One adjoint matrix-vector product data point is used per forward model operator data point. & \eqref{eqn:loss-adjvec} & $1\text{E-}3$\\  \hline
   
   \mycell{{\tt Indep}\\ ($2$ Networks)} & Forward model operator data points and corresponding full adjoint matrices used to train two separate networks. {\tt IndepFwd} considers only forward model operator data point mismatch, and {\tt IndepAdj} considers only adjoint model data point mismatch. & 
   \begin{tabular}{c} \eqref{eqn:loss-standard} ({\tt IndepFwd}), \\ \eqref{LossIndepAdj} ({\tt IndepAdj}) \end{tabular} & None  \\  \hline
   
     \mycell{{\tt IndepVec}\\ ($2$ Networks)} & Forward model operator data points with corresponding adjoint data provided as matrix-vector products used to train two separate networks. {\tt IndepVecFwd} considers only forward model operator data point mismatch, and {\tt IndepVecAdj} considers only adjoint model data point mismatch. & 
   \begin{tabular}{c} \eqref{eqn:loss-standard} ({\tt IndepVecFwd}), \\ \eqref{LossIndepAdjVec} ({\tt IndepVecAdj}) \end{tabular} & None\\  \hline
   
    \mycell{{\tt Standard} \\ ($1$ Network)} & Forward model operator data points only used to train one network. & \eqref{eqn:loss-standard} & None  \\  \hline
    \end{tabularx}
\end{table}

\begin{table}
    \centering
    \caption{Summary of neural network adjoint-vector based surrogate model construction methods and a baseline method used for comparison. {\tt Random} in \Cref{table-adjvec-methods} is the same network as {\tt AdjVec} in  \Cref{table-surrogate-methods}.}
    \begin{tabularx}{\textwidth}{| l | X | X | l |}
    \hline
   Method & Description & Vectors Used For Adjoint-Vector Products & $\alpha$\\ \hline
    \mycell{{\tt Standard} \\ ($1$ Network)} & Forward model operator data points only used to train one network. This is the same network as {\tt Standard} in \ref{sec:NN-surrogate-methods}. & None & None \\  \hline
   \mycell{{\tt Lagrange} \\ ($1$ Network)} & Adjoint-vector products computed with vectors approximating the Lagrange multiplier terms from \Cref{sec:theory}.  & \eqref{eqn:adjoint-vector-computable-form} & $1\text{E-}5$\\  \hline
   \mycell{{\tt Random} \\ ($1$ Network)} & Adjoint-vector products computed by multiplying the adjoint matrix with standard Gaussian random vectors. This is the same network as {\tt AdjVec} in \Cref{sec:NN-surrogate-methods}. & Standard Gaussian random vectors & $1\text{E-}3$ \\  \hline
   \mycell{{\tt RandCol} \\ ($1$ Network)} & Adjoint-vector products computed by multiplying the adjoint matrix with randomly selected standard basis vectors. & Standard basis vectors selected randomly from a uniform distribution. & $5\text{E-}2$\\  \hline
    \end{tabularx}
    \label{table-adjvec-methods}
\end{table}

{\tt Adj}, {\tt AdjVec}, and {\tt Standard} models use networks of the form
\begin{equation}\label{eqn:nneqn}
    \nnm(\uin;\theta) = \mathbf{W}_{2}\tanh(\mathbf{W}_{1}\uin + \mathbf{b}_{1}) + 
    \mathbf{b}_{2},
\end{equation}
where $\theta$ is the network's parameter vector which specifies $\mathbf{W}_{1}$, $\mathbf{W}_{2}$,
$\mathbf{b}_{1}$, and $\mathbf{b}_{2}$.   In this experiment, the dimensions specified by our test problem are $\Nsub{in} = \Nsub{out} = 3$.
 In this work we use two layers ($k = 2$), a hidden dimension $\Nsub{hidden} = 25$. This specifies the sizes of all network weight parameter matrices and vectors: 
 \begin{subequations}
\begin{align}\label{eqn:2layer-NN}
    \mathbf{W}_{1} &\in \Re^{\Nsub{hidden} \times \Nstate}, & \mathbf{W}_{2} & \in \Re^{\Nstate \times \,\Nsub{hidden}}, \\
    \mathbf{b}_{1} &\in \Re^{\Nsub{hidden}}, & \mathbf{b}_{2} & \in \Re^{\Nstate}.
\end{align}
\end{subequations}
These are used in place of the high-fidelity model in the solution of the 4D-Var optimization problem.

{\tt Indep} and {\tt IndepVec} use two networks each: one  network used for computing the model operator operations in \Cref{eqn:4dvar-pert}, denoted {\tt IndepFwd} and {\tt IndepVecFwd} respectively in \Cref{table-surrogate-methods}, and a second set of models, {\tt IndepAdj} and {\tt IndepVecAdj}, used to approximate the high fidelity model operator adjoint in the gradient calculations. {\tt IndepFwd} and {\tt IndepVecFwd} are networks of the form \Cref{eqn:nneqn}. For direct comparability of the methods, we have introduced a non-standard neural architecture for {\tt IndepAdj} and {\tt IndepVecAdj}. The derivative of \Cref{eqn:nneqn} with respect to its input is
\begin{equation}\label{eqn:nntlm}
    \nnmtlm(\uin;\theta) = \mathbf{W}_{2}\, \text{diag}\left(\text{sech}^{2}(\mathbf{W}_{1}\uin + \mathbf{b}_{1})\right)\, \mathbf{W}_{1}. 
\end{equation}
We use networks of the form \Cref{eqn:nntlm} for both {\tt IndepAdj} and {\tt IndepVecAdj}. This ensures that the single-network and dual-network methods have exactly the same expressivity while learning the network's derivative function.

All adjoint-vector product methods, explored \Cref{sec:adjv-methods}, use the standard two layer network architecture described in \cref{eqn:nneqn}.

\subsubsection{Discussion of Architecture Choice}
 While the network described in \Cref{eqn:nneqn} is a very shallow network, the universal approximation theorem for neural networks \cite{HORNIK1991251} guarantees that a network of this structure can approximate any continuous function for sufficiently large hidden dimension size. Here we chose a small, shallow  network to represent a computationally feasible surrogate of a larger, more expensive system. As shown in \Cref{table-4dvar,table-time-average}, this choice of architecture, with optimal training and data, is capable of producing a surrogate model giving 4D-Var solutions with errors that are within $1.3$\% of exact 4D-Var errors, but obtained in an order of magnitude smaller computational time. The small network is an appropriate surrogate for the Lorenz'63 test problem.
 
Because the neural network is used with a 4D-Var optimization procedure (BFGS) that estimates the loss function's second derivative, the authors believe that the choice of smooth $\tanh$ activation \cite{borreljensen2021physicsinformed} is preferable to ReLU, which has zero second derivative everywhere  it is defined.

\subsubsection{Training Data}
\label{subsec-trainingdata}
Three groups of data are collected and used in the training process. Forward model data $\{(\danx_{t_{i}},\Model_{t_{i}, t_{i+1}}(\danx_{t_{i}}))\}_{i = 1}^{\Nsub{data}}$ is used in training {\tt Standard}, {\tt Adj}, {\tt AdjVec}, and {\tt IndepFwd}. Adjoint data $\{\Madjt_{t_{i}, t_{i} + 1} \}_{i = 1}^{\Nsub{data}}$ is used in training {\tt Adj} and  ${\tt IndepAdj}$. Adjoint-vector product data of the form $\{\Madjt_{t_{i}, t_{i} + 1}\mathbf{v}_{i} \}_{i = 1}^{\Nsub{data}}$ is used in training {\tt AdjVec} and {\tt IndepVecAdj}. The adjoint-vector product methods, analyzed in \Cref{sec:adjv-methods}, also use data of the same form used by {\tt AdjVec}. The vectors used by each of these method are summarized in  \Cref{table-adjvec-methods}.

We use explicit fourth order Runge-Kutta to collect forward model data, and its corresponding discrete adjoint method to generate adjoint data with the same settings described in \Cref{sec:ground-truth-generation}. Data is generated by one extended integration of $\Nsub{data} = 500$ intervals of length $\Delta T$ from a randomly chosen initial system state
\begin{equation}\label {eqn:training-initial-condition}
 \mathbf{x}_0  = \begin{bmatrix}
   -5.9448\\
    -5.6587 \\
    24.4367 
    \end{bmatrix} + \boldsymbol{\varepsilon}, \qquad \boldsymbol{\varepsilon} \sim \mathcal{N}(0_{3 \times 1}, 5\cdot\mathbf{I}_{3 \times 3}).
\end{equation}
Note that trajectories that provide training data start from initial conditions \eqref{eqn:training-initial-condition} different from the reference values \eqref{eqn:fixed-l63-initial-condition}.
The training data is calculated separately for each method, i.e., each method is trained on data obtained from trajectories started from independently sampled initial conditions \eqref {eqn:training-initial-condition}. Adjoint-vector data is computed using the adjoint matrices from the calculated trajectory multiplied with the vectors chosen for each method, which are summarized in \Cref{table-adjvec-methods}.

\subsubsection{Network Training Algorithm and Settings}
Training for all models is done with Adam \cite{Kingma2014}, using $100$ batches of size $5$ per epoch, and $200$ epochs. Learning rates are scheduled  log-uniformly over training epochs from a maximum of $1e-2$ to a minimum of $1e-5$. 
\subsubsection{Regularization Parameter Selection}
Loss functions and weight coefficient $\alpha$ values used for each method are summarized in \Cref{table-surrogate-methods} and \Cref{table-adjvec-methods}. The $\alpha$ values were determined by independent parameter sweeps for each training method with the goal of optimal 4D-Var spatiotemporal RMSE.

\subsubsection{Measurement of Model Generalization Performance}\label{sec:generalization-setup}
Test data used in the forward model generalization error test is collected as follows. The initial point is \Cref {eqn:training-initial-condition}, randomly perturbed by noise sampled from $\mathcal{N}(0, \mathbf\danB_{0})$. 
One integration of $10,000$ time steps of length $\Delta T$ is performed, with system state again perturbed by noise sampled from $\mathcal{N}(0, \mathbf\danB_{0})$ every $500$ steps to generate a large quantity of test data outside of the training data set. Forward and adjoint information is collected. The generalization test represents the accuracy of the model on data not used in the training process removed from the context of solution to the 4D-Var problem. Forward model performance is calculated using the formula
\begin{equation}\label{eqn:gen-rmse}
\text{RMSE} = \sqrt{\sum_{i=1}^{\Nsub{data}}
\frac{\norm{\mathcal{N}(\danx_{t_{i}};\theta) - \Model_{t_{i}, t_{i+1}}(\danx_{t_{i}})}^{2}_{2}}{\Nstate\cdot\Nsub{data}}
}.
\end{equation}
Adjoint model performance is calculated using
\begin{equation}\label{eqn:adjoint-rmse}
    \text{RMSE} = \sqrt{\sum_{i=1}^{\Nsub{data}} \frac{\norm{ \nnmtlm^{T}(\danx_{t_{i}};\theta) - \Madjt_{t_{i}, t_{i+1}}(\danx_{t_{i}})}_F^2}{\Nsub{data}\cdot\Nstate^{2}}}.
\end{equation}

\section{Results and Discussion}\label{sec:results-and-discussion}
\subsection{Comparison Between Neural Network Surrogate Methods}\label{sec:NN-surrogate-methods}
In this section we report results for the methods summarized in \Cref{table-surrogate-methods}.

\subsubsection{Accuracy of 4D-Var Solutions Using Different Surrogates}\label{sec:4dvar-accuracy}
\begin{table}
    \centering
    \caption{Mean spatiotemporal RMSE and RMSE standard deviations of 4D-Var analyses obtained with different surrogate models. Results averaged over $15$ independent trials.}
    \label{table-4dvar}
    \begin{tabularx}{\textwidth}{| X | X | X | X | X | X | X | X |}
    \hline
   Method & {\tt Forecast} & {\tt Exact} & {\tt Adj} & {\tt AdjVec} & {\tt Standard} & {\tt Indep} & {\tt IndepVec} \\ \hline
    Analysis RMSE & $12.08 \pm 0.37$ & $0.83 \pm 0.03$  & $0.84 \pm 0.03$ & $0.97 \pm 0.04$ & $1.08 \pm 0.03$ & $0.97 \pm 0.08$ & $1.21 \pm 0.30$\\  \hline
    \end{tabularx}
\end{table}
\Cref{table-4dvar} reports the accuracy of the 4D-Var solutions obtained with each surrogate model discussed. Using two-sided paired T-tests, 4D-Var spatiotemporal RMSEs differ significantly from {\tt Standard} at $p = .95$. Under the same two-sided paired T-Tests, the only pairs of methods that do not differ significantly in terms of spatiotemporal RMSE at $p = 0.95$ are {\tt Standard} and {\tt IndepVec}, and {\tt AdjVec} and {\tt Indep}.

Except for {\tt IndepVec}, all methods that incorporate derivative information ({\tt Adj}, {\tt AdjVec}, and {\tt Indep}) provide better solutions than {\tt Standard}. \sandu{??} provides very accurate analyses, nearly matching {\tt Exact}. However, incorporating full adjoint matrix data into the training process is computationally infeasible in many settings. The experiment indicates the possibility of excellent accuracy benefits in contexts where it is feasible. 

While {\tt Indep} and {\tt IndepVec} provide more accurate modeling of the adjoint model dynamics than {\tt Adj}, as seen in \Cref{table-adjoint-generalization}, the use of independent networks poses problems for numerical optimization routines. Because the networks are independent, the surrogate model gradients computed using the adjoint networks ({\tt IndepAdj} and {\tt IndepVecAdj}) need not be descent directions for the for the surrogate 4D-Var cost functions computed using the forward networks ({\tt IndepFwd} and {\tt IndepVecFwd}). This behavior can be observed in \Cref{opttime-image}, where {\tt Indep} spends a significant amount of time stagnating before the optimization concludes. In this case, the surrogate gradient computed using {\tt IndepAdj} was an ascent direction for the surrogate cost function computed using {\tt IndepFwd}. In addition, training the adjoint network used in {\tt Indep}, {\tt IndepAdj}, poses the same computational difficulties faced by {\tt Adj}, since it is trained using full adjoint matrices, suggesting that training one single network provides a better use of computational time and data when training with full adjoint matrices is feasible.

\subsubsection{Forward Model Generalization Performance}
\begin{table}
    \centering
    \caption{Forward model generalization performance. Average and standard deviations of $15$ test data set realizations.}
    \label{table-generalization}
    \begin{tabular}{| l | c | c | c |}
    \hline
   Method & {\tt Standard} & {\tt Adj} & {\tt AdjVec} \\ \hline
    RMSE & $0.48 \pm 0.05$ & $0.13 \pm 0.06$ & $0.44 \pm 0.05$  \\ \hline
    \end{tabular}
\end{table}

\Cref{table-generalization} reports the generalization performance of the {\tt Standard}, {\tt Adj}, and {\tt AdjVec} surrogates on unseen test data. Test data is generated as described in Section \ref{subsec-trainingdata}. RMSE is calculated using \Cref{eqn:gen-rmse}. The table reports average RMSE from $15$ independent test data set generations. Because the forward models of {\tt Indep} and {\tt IndepVec} are the same network as {\tt Standard}, they are not tested. Paired, two-sided T-Tests showed that all pairwise differences in test set RMSE were significant at $p = 0.95$.

The incorporation of adjoint information into the training process appears to also improve the accuracy of the forward model, as seen in lower test error for {\tt Adj} and {\tt AdjVec} compared to {\tt Standard}. Incorporating full adjoint matrices as in {\tt Adj}'s loss function \eqref{eqn:loss-adjoint} leads to better results than incorporating only adjoint-vector products as in {\tt AdjVec}'s loss function \eqref{eqn:loss-adjvec}, but improved accuracy is obtained in both cases.

\subsubsection{Adjoint Model Generalization Performance}
\begin{table}
\centering
    \caption{Adjoint model generalization performance. Average and standard deviations of $15$ test data set realizations.}
    \begin{tabular}{| l | c | c | c | c | c |}
    \hline
   Method & {\tt Standard} & {\tt Adj} & {\tt AdjVec} & {\tt Indep} & {\tt IndepVec} \\ \hline
    RMSE & $0.97 \pm  0.01$ & $0.06 \pm 0.02$ & $1.12 \pm 0.00$ & $0.02 \pm 0.00$ & $0.05 \pm 0.00$ \\ \hline
    \end{tabular}
    \label{table-adjoint-generalization}
\end{table}

\Cref{table-adjoint-generalization} shows adjoint model generalization performance compared between {\tt Standard}, {\tt Adj}, {\tt AdjVec}, {\tt Indep}, and {\tt IndepVec} on unseen training set data. The adjoint for each network is calculated either by differentiating the network at the intial condition given by the test data set in the case of the single-network methods, or taking the output of the adjoint network applied to the initial condition for the two-network methods, and then comparing to the corresponding test data set adjoint value. Test data is generated as described in Section \ref{sec:generalization-setup}. RMSE is calculated using the formula \Cref{eqn:adjoint-rmse}. The table reports average RMSE from fifteen independent test data set generations. 

Three of the four methods that incorporate adjoint information ({\tt Adj}, {\tt Indep}, and {\tt IndepVec}) demonstrate better modeling of the adjoint dynamics than {\tt Standard}. Surprisingly, {\tt AdjVec} demonstrates worse adjoint test performance than {\tt Standard}. The incorporation of full adjoint matrices as in {\tt Adj}'s loss function \eqref{eqn:loss-adjoint} provides more benefit than incorporating only adjoint-vector products in {\tt AdjVec}. {\tt Indep} shows significantly improved accuracy in matching the full model's adjoint dynamics, and similarly shows greater improvement than {\tt IndepVec}. {\tt Adj} does not demonstrate better accuracy than {\tt Indep} or {\tt IndepVec}, indicating that incorporating forward model data into the training process does not improve modeling of the adjoint dynamics.  Paired, two-sided T-Tests showed that all pairwise differences in test set RMSE were significant at $p = 0.95$.

\subsubsection{4D-Var Optimization Convergence and Timing}
\label{sec:4dvar-convergence}
\begin{table}
\centering
    \caption{Average wall times and standard deviations for all surrogate optimizations during the sequential 4D-Var run.}
    \begin{tabularx}{\textwidth}{| X | X | X | X | X | X | X |}
    \hline
   Method & {\tt Exact} & {\tt Adj} & {\tt AdjVec} & {\tt Standard} & {\tt Indep} & {\tt IndepVec}\\ \hline
    Wall Time (s)  & $1.70\text{E-2} \pm 4.26\text{E-3}$  & $1.73\text{E-3} \pm 4.39\text{E-4}$ & $3.61\text{E-3} \pm 1.10\text{E-3}$ & $3.58\text{E-3} \pm 1.17\text{E-3}$ & $2.93\text{E-3} \pm 9.53\text{E-4}$ & $2.99\text{E-3} \pm 1.06\text{E-3}$ \\  \hline
    \end{tabularx}
    \label{table-time-average}
\end{table}

\begin{figure}[h]
\begin{center}
\includegraphics[width=.65\linewidth]{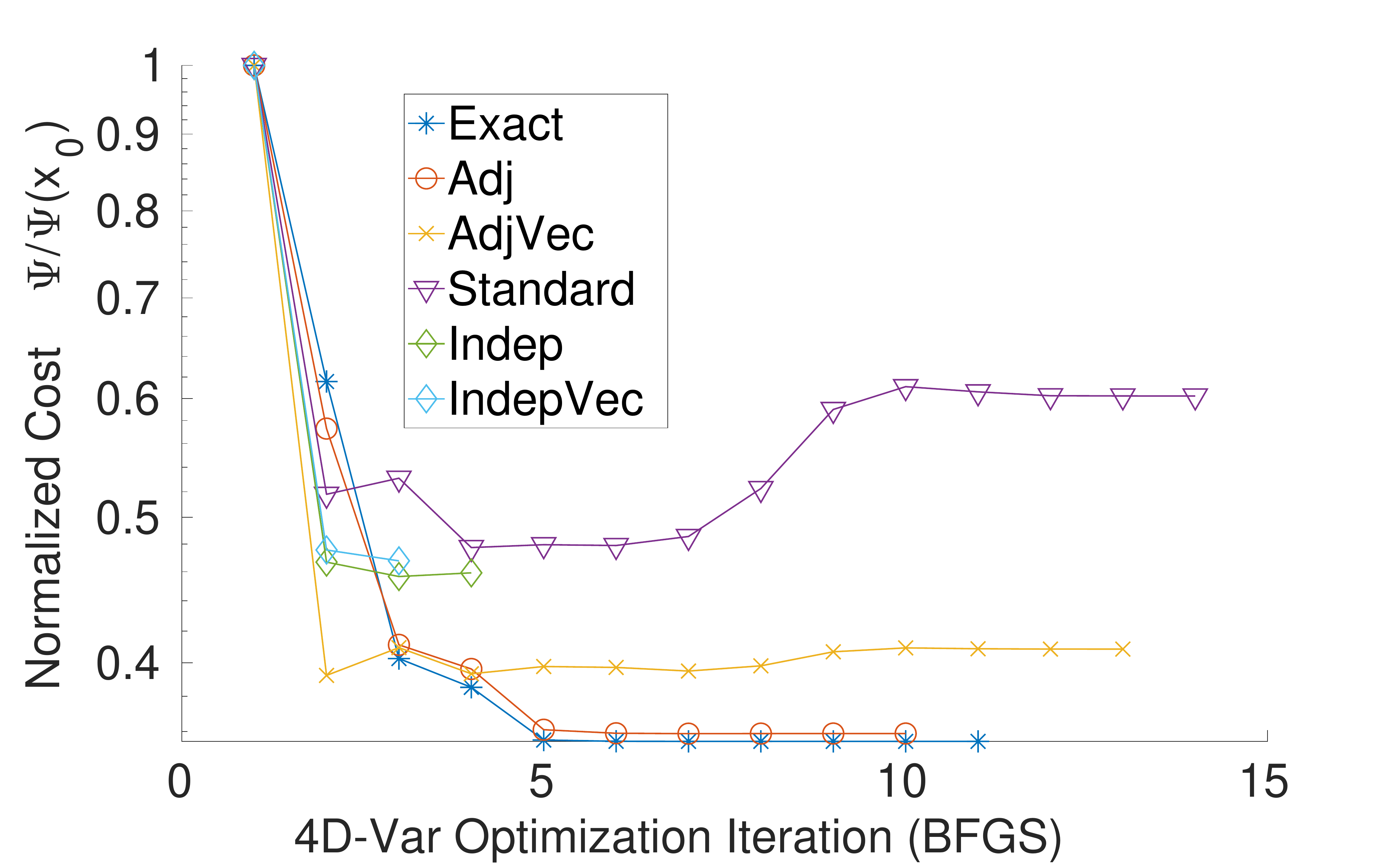}
\end{center}
\caption{ Cost values of the high-fidelity 4D-Var cost function \eqref{eqn:4dvar} during optimization of the surrogate 4D-Var problem with different surrogate models. The cost is normalized by its initial value. All optimizations are initialized with the same starting point, and  {\tt fminunc}'s default stopping criteria are used.}
\label{optiters-image}
\end{figure}

\begin{figure}[h]
\begin{center}
\includegraphics[width=.65\linewidth]{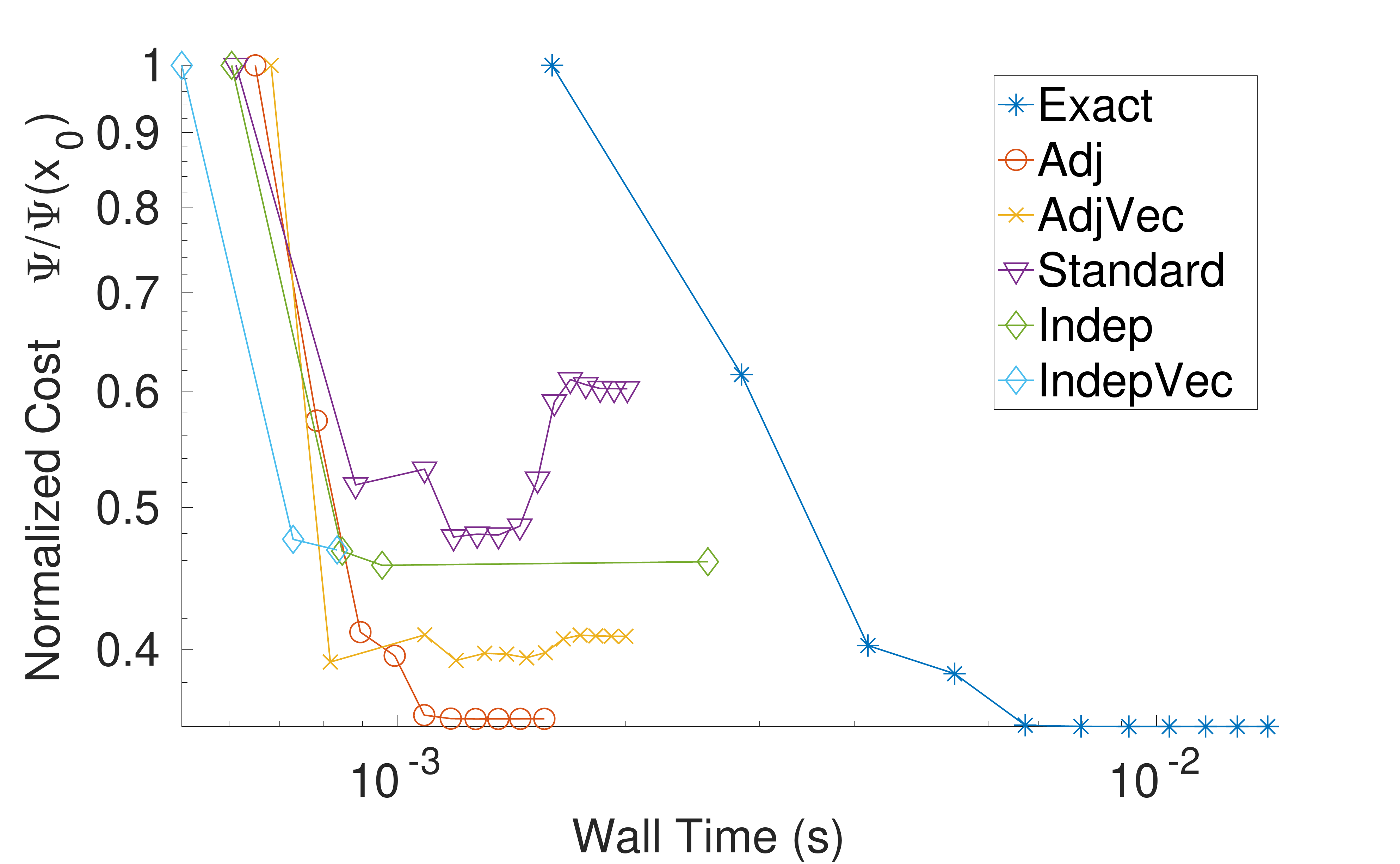}
\end{center}
\caption{Decrease of the high-fidelity 4D-Var cost function values \eqref{eqn:4dvar} during optimization of the surrogate 4D-Var problems against wall-clock computational time. The differences in starting time are due to time differences in the single function evaluation used by MATLAB's {\tt fminunc} at iteration $0$.}
\label{opttime-image}
\end{figure}

As shown in \Cref{table-time-average}, all neural network-based methods take significantly less time to reach convergence on the 4D-Var optimization problem \cref{eqn:4dvar} than {\tt Exact}. All optimizations take place during a sequential 4D-Var run of $550$ time steps in exactly the same setup as described in \Cref{sec:4dvar-experiment}. Every surrogate method's time differed significantly from {\tt Exact} under two-sided paired T-tests at $p = 0.95$.

Figures \ref{optiters-image} and \ref{opttime-image} demonstrate the behavior of the 4D-Var optimization using various surrogates. All optimizations start from a fixed, common initial guess. As noted in \Cref{sec:4dvar-accuracy}, the 4D-Var gradients produced by {\tt Indep} need not be descent directions to the optimization problem. At the optimization routine's stopping point, ${\tt Indep}$'s gradient computed using $\nnm_{\rm IndepAdj}$ was in fact an ascent direction for the surrogate cost function computed using $\nnm_{\rm IndepFwd}$, causing the optimization to terminate after a fruitless line search. 

\subsubsection{Loss During the Training Process}
\label{sec:loss-during-training}
\begin{figure}[h]
\begin{center}
\includegraphics[width=.65\linewidth]{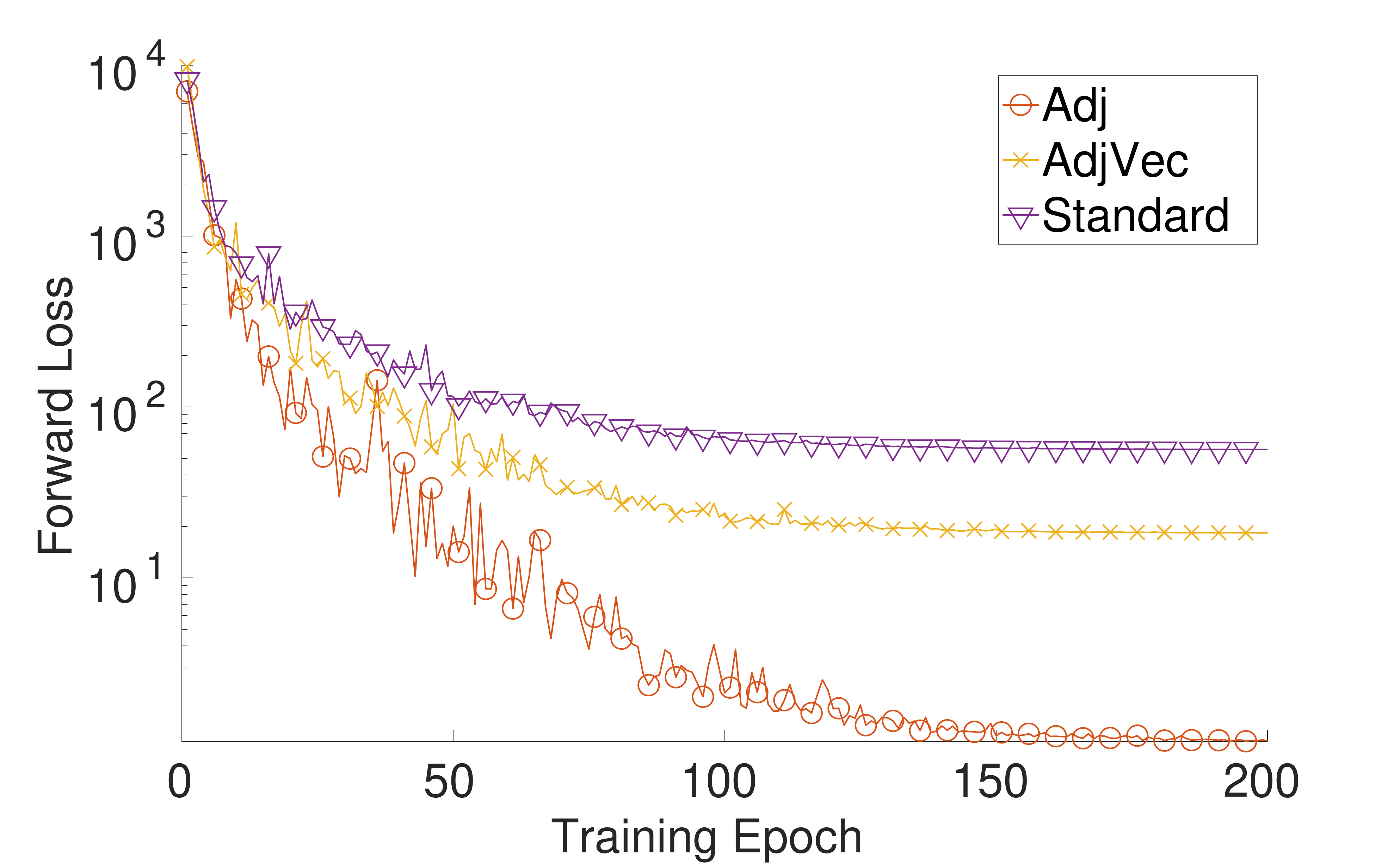} 
\end{center}
\caption{Forward loss of each network during the training process.} \label{fig:fwd-training-costs}
\end{figure}

\begin{figure}[h]
\begin{center}
\includegraphics[width=.65\linewidth]{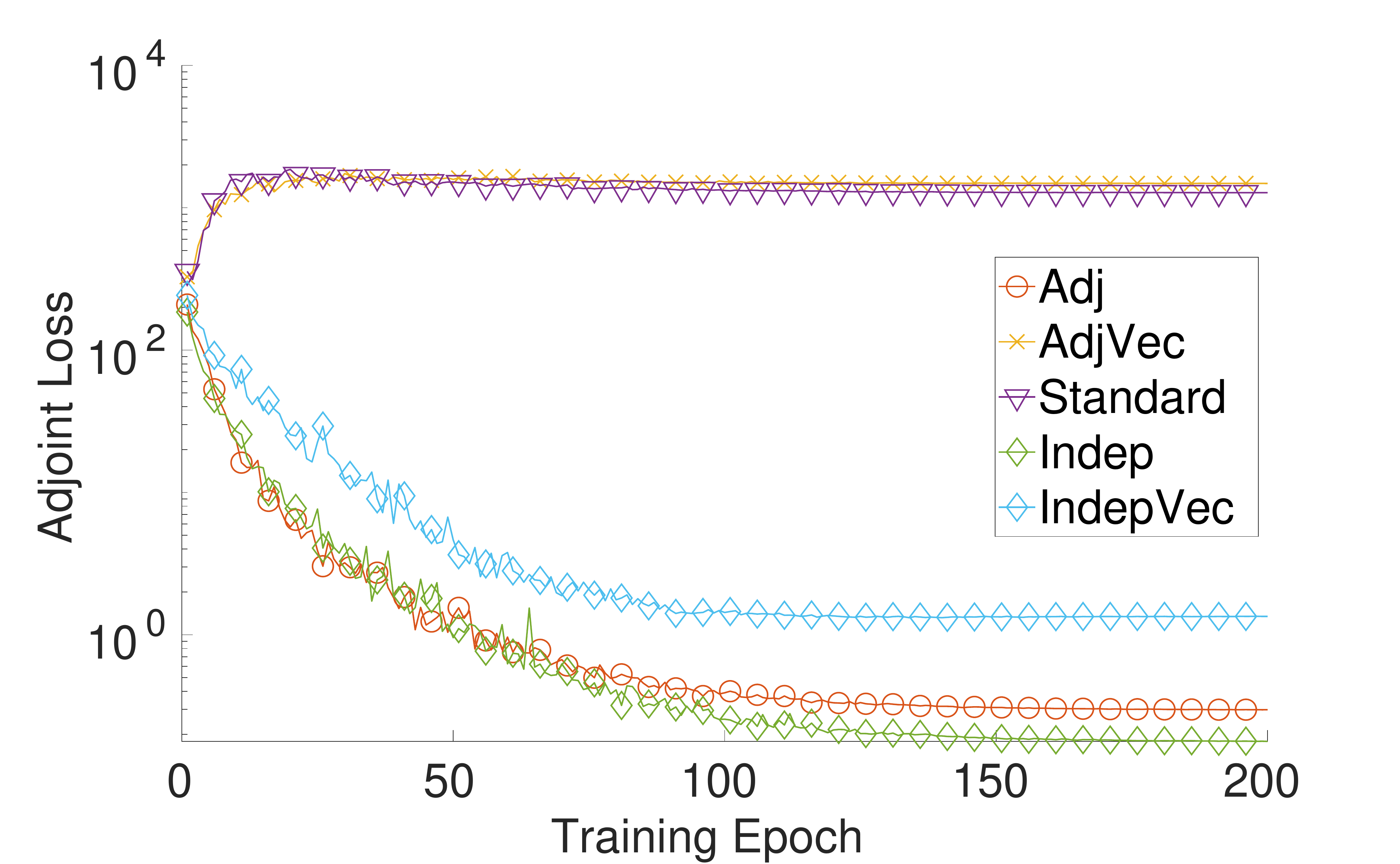} 
\end{center}
\caption{Adjoint loss of each network during the training process.}\label{fig:adj-training-costs}
\end{figure}

Figures \ref{fig:fwd-training-costs} and \ref{fig:adj-training-costs} show separately the decrease of the forward training loss and of the adjoint loss during the training process.  The forward loss is exactly $\loss{Standard}$ in $\eqref{eqn:loss-standard}$, where the summation is over the entire training data set, calculated at the end of each training epoch. Similarly, adjoint mismatch is calculated using \eqref{LossIndepAdj}, where the summation also takes place over the entire training data set at the end of each epoch. As in the generalization tests, the adjoint for each network is calculated either by differentiating the network at the intial condition given by the test data set in the case of the single-network methods, or taking the output of the adjoint network applied to the initial condition for the two-network methods, and then comparing to the corresponding test data set adjoint value.

Inclusion of full adjoint matrix information in {\tt Adj} greatly speeds up, in terms of training epochs and optimization iterations, convergence to high training set accuracy in Figure \ref{fig:fwd-training-costs}. {\tt AdjVec} also has faster forward mismatch convergence. This suggests that adding adjoint information can lead to better forward models in fewer training epochs. In Figure \ref{fig:adj-training-costs}, we see that {\tt Standard} and {\tt AdjVec} do not learn to more accurately model the adjoint data, in terms of the training data, during the training process. By contrast, {\tt Adj}, {\tt AdjVec}, {\tt Indep}, and {\tt IndepVec} do. The additional computation of the neural network's adjoint, adjoint-vector product, or the evaluation of a separate neural network should cause an increase in wall time per cost function and gradient evaluation during hte training process. The methods did not differ from each other in terms of wall time in our experiments. The authors believe this result is a spurious and non-representative result, likely occurring due to the small scale of our tests.

\subsection{Comparison Between Adjoint-Vector Product Methods}\label{sec:adjv-methods}
This section considers only the methods summarized in \Cref{table-adjvec-methods}. 

In \Cref{sec:theory}, we argued that accurate solution of the surrogate problem requires that the surrogate model accurately match the forward high-fidelity model dynamics, as well as match the high-fidelity model's adjoint vector products with Lagrange multipliers derived from the constrained problem's optimality conditions. As noted in \Cref{sec:theory}, these vectors can only be approximated during the training process, as they are implicitly defined in terms of the trained surrogate model itself. In \Cref{sec:vector-choice}, we described several computable approximations to these vectors, one of which \eqref{eqn:adjoint-vector-computable-form} we used to train the {\tt AdjVec} method in \Cref{sec:NN-surrogate-methods}. In this section, we compare the performance of several adjoint-vector product methods, trained as described in \Cref{subsec-adjv-training}, using the vector choices described in \Cref{sec:vector-choice}.

\subsubsection{Accuracy of 4D-Var Solutions}\label{sec:adjv-4dvar-accuracy}
\begin{table}
    \centering
    \caption{Mean spatiotemporal RMSE and standard deviations of 4D-Var analyses obtained with different adjoint-vector surrogate models. Results averaged over $15$ independent trials.}
    \begin{tabularx}{\textwidth}{| X | X | X | X | X |}
    \hline
   Method & {\tt Standard} & {\tt Lagrange} & {\tt Random} & {\tt RandCol} \\ \hline
    Analysis RMSE & $1.08 \pm 0.03$ & $1.03 \pm 0.06$  & $0.97 \pm 0.04$ & $1.04 \pm 0.04$  \\  \hline
    \end{tabularx}
    \label{table-adjvec4dvar}
\end{table}

The average 4D-Var solution RMSEs from a sequential 4D-Var run is reported in \Cref{sec:adjv-4dvar-accuracy}. Under paired two-sided T-Tests, all pairs of method accuracies were significantly different at $p = 0.95$ except for {\tt RandCol} and {\tt Lagrange}.

{\tt Random} provides the best average 4D-Var solution accuracy compared to all other adjoint-vector product methods. {\tt Lagrange}, {\tt Random}, and {\tt Half} all show lower average RMSE than {\tt Standard}, and therefore offer improvements over it.

In \eqref{eqn:adjoint-vector-computable-form}, our derivation for the vectors used in {\tt Lagrange}, we see that the vectors are multiplied on the left by the transpose of the linearized observation operator $\HH_{t_{i} + 1}^T$. For our test problem, outlined in \Cref{sec:4dvar-experiment}, this matrix has zeros in its second column, meaning that the resulting adjoint-vector product has $0$ in its second coordinate. This means that {\tt Lagrange} incorporates no information about the second column of the adjoint matrix into the training process. This may offer some intuitive explanation for how {\tt Random} may gain more from adjoint-vector product training, however, {\tt RandCol} contains information about all $3$ coordinates, and does not statistically differ from {\tt Lagrange} in terms of 4D-Var solution accuracy. Training to preserve adjoint-vector products with random vectors may retain some special benefit. It is interesting to note that, as we will see in \Cref{table-adjvec-generalization}, {\tt Random} does not differ significantly from {\tt RandCol} in terms of forward model generalization, and is actually shown to be significantly worse than {\tt Standard}, {\tt Lagrange}, and {\tt RandCol} in terms of adjoint generalization performance, as seen in \Cref{table-adjvec-adjoint-generalization}. This illustrates the importance of testing the proposed training methods in the context of the application of interest.

\subsubsection{Forward Model Generalization Performance}
\begin{table}
    \centering
    \caption{Forward model generalization performance. Average and standard deviations of $15$ test data set realizations.}
    \begin{tabular}{| l | c | c | c | c |}
    \hline
   Method & {\tt Standard} & {\tt Lagrange} & {\tt Random} & {\tt RandCol} \\ \hline
    RMSE & $0.48 \pm 0.05$ & $0.39 \pm 0.02$ & $0.43 \pm 0.03$  & $0.43 \pm 0.03$\\ \hline
    \end{tabular}
    \label{table-adjvec-generalization}
\end{table}

\Cref{table-generalization} reports the generalization performance of the {\tt Standard}, {\tt Lagrange}, {\tt Random}, and {\tt RandCol} surrogates on unseen test data. Test data is generated as described in Section \ref{subsec-trainingdata}. RMSE is calculated using \Cref{eqn:gen-rmse}. The table reports average RMSE from $15$ independent test data set generations. Paired, two-sided T-Tests showed that all pairwise differences in test set RMSE were significant at $p = 0.95$.

All methods of incorporating adjoint-vector data lead to improved forward models compared to {\tt Standard}, but {\tt Lagrange} provides the most benefit. The best model in terms of forward model generalization performance in \Cref{table-adjvec-generalization} differs from the best model in terms of 4D-Var RMSE in \Cref{table-adjvec4dvar}. The quality of the 4D-Var solution cannot be explained simply by improved modeling of the forward dynamics. 

\subsubsection{Adjoint Model Generalization Performance}
\begin{table}
\centering
    \caption{Adjoint model generalization performance on $10,000$ out of dataset adjoint test points for adjoint-vector product methods. Generalization performance of the models increases with the incorporation of adjoint information into the training data. Mean total RMSE and standard deviations of $10$ test data realizations.}
    \begin{tabular}{| l | c | c | c | c |}
    \hline
   Method & {\tt Standard} & {\tt Lagrange} & {\tt Random} & {\tt RandCol} \\ \hline
    RMSE & $0.97 \pm 0.01$ & $1.05 \pm 0.00$ & $1.12 \pm 0.00$  & $0.74 \pm 0.01$\\ \hline
    \end{tabular}
    \label{table-adjvec-adjoint-generalization}
\end{table}

\Cref{table-adjoint-generalization} shows adjoint model generalization performance compared between {\tt Standard}, {\tt Lagrange}, {\tt Random}, and {\tt RandCol} on unseen training set data. The adjoint for each network is calculated either by differentiating the network at the intial condition given by the test data set. Test data is generated as described in Section \ref{sec:generalization-setup}. RMSE is calculated using the formula \Cref{eqn:adjoint-rmse}. The table reports average RMSE from fifteen independent test data set generations.

Adjoint model generalization performance is given in \Cref{table-adjvec-adjoint-generalization}. All pairs of methods differed significantly under paired, two-sided T-Tests at $p = 0.95$. It is interesting to note that only {\tt RandCol} provides better adjoint matrix generalization than {\tt Standard}. {\tt Random} and {\tt Lagrange} provide worse modeling of the system adjoints. A comparison with 4D-Var RMSE in \Cref{table-4dvar} shows that improvement in the 4D-Var solution accuracy cannot be explained exclusively by improved adjoint modeling.

\subsubsection{Loss During the Training Process}
\begin{figure}[h]
\begin{center}
\includegraphics[width=.65\linewidth]{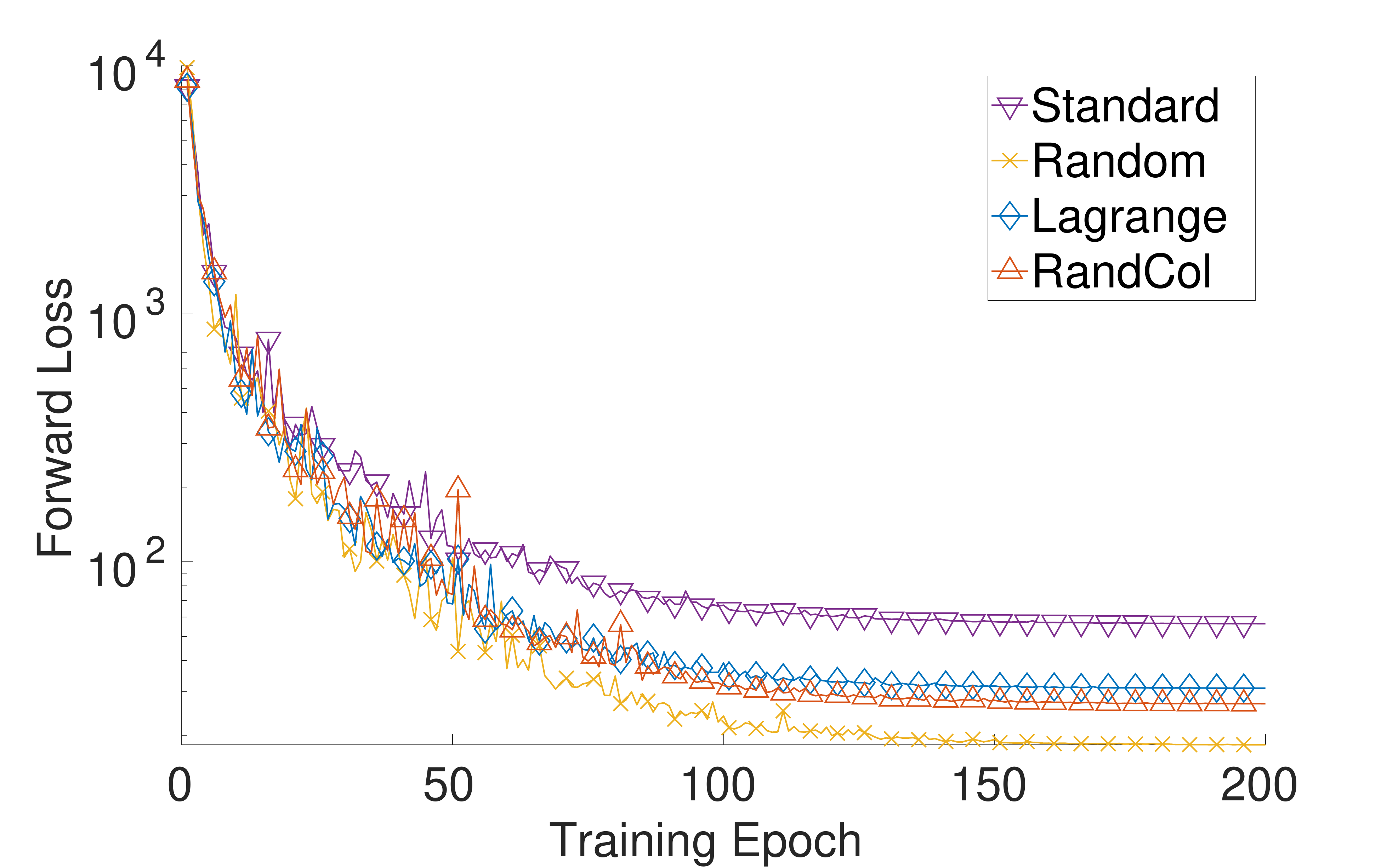} 
\end{center}
\caption{Forward loss of each network during the training process.} \label{fig:adjv-fwd-training-costs}
\end{figure}

\begin{figure}[h]
\begin{center}
\includegraphics[width=.65\linewidth]{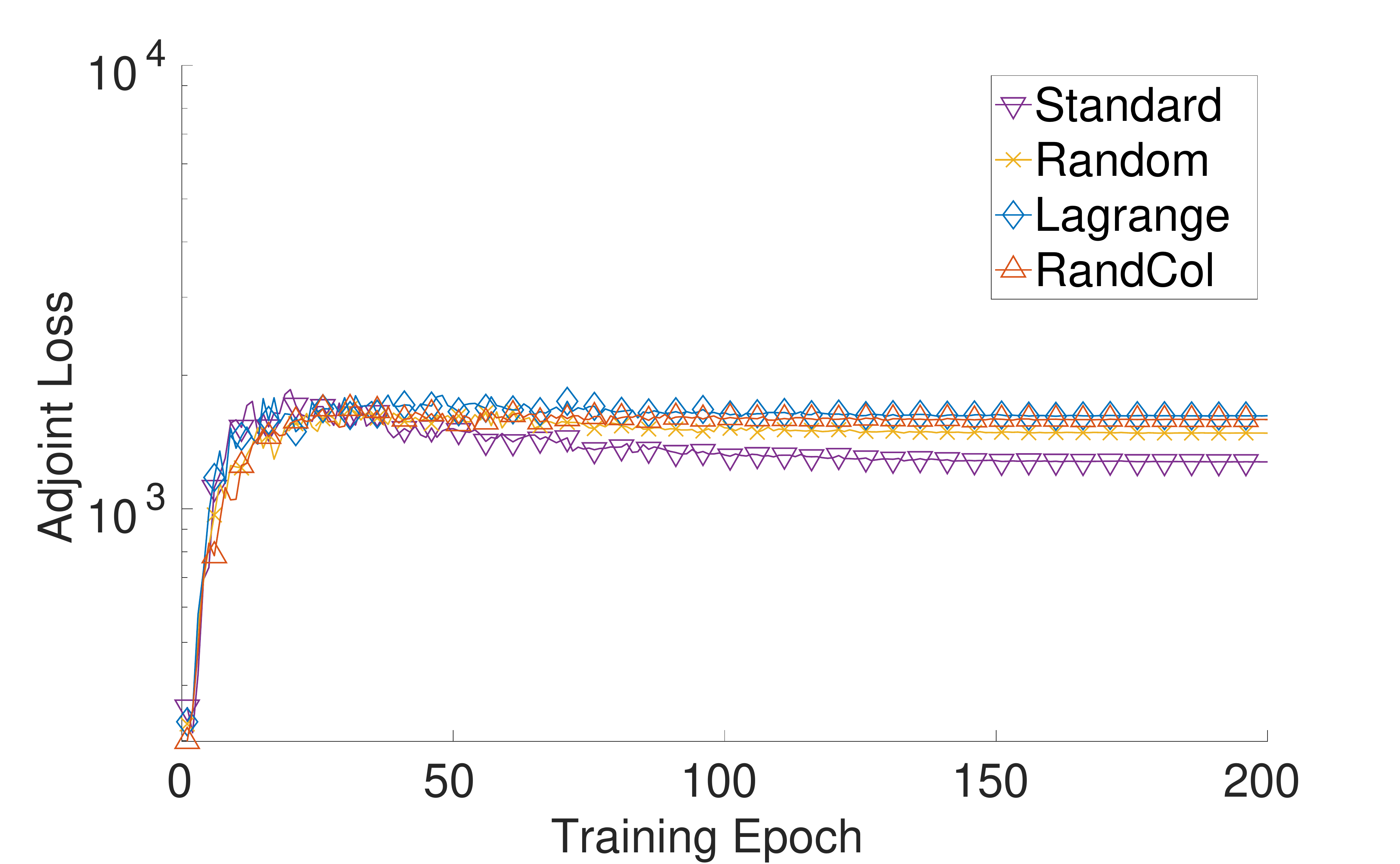} 
\end{center}
\caption{Adjoint loss of each network during the training process.}\label{fig:adjv-adj-training-costs}
\end{figure}

The training loss curves for {\tt Random}, {\tt Lagrange}, and {\tt RandCol} are shown in \Cref{fig:adjv-fwd-training-costs} and \Cref{fig:adjv-adj-training-costs}. As in the generalization tests, the adjoint for each network is calculated either by differentiating the network at the intial condition given by the test data set in the case of the single-network methods, or taking the output of the adjoint network applied to the initial condition for the two-network methods, and then comparing to the corresponding test data set adjoint value.

The training loss curves roughly match the behavior observed in \Cref{sec:loss-during-training} and Table \ref{table-adjvec-generalization}. {\tt Random}, {\tt Lagrange}, and {\tt RandCol} achieve better forward loss on the training data, while underperforming {\tt Standard} in terms of adjoint loss.

\section{Conclusions and Future Directions}
\label{sec-conclusions}
This work constructs science-guided neural network surrogates of dynamical systems for use in variational data assimilation. The new surrogates match not only the dynamics, but also the derivatives of the original dynamical system. Specifically, we consider several training methodologies for the neural networks that incorporate model adjoint information into the loss function. Our results suggest that incorporating of derivative information in the surrogate construction results in significantly improved solutions to the 4D-Var problem. The quality of the forward model, as measured by its generalization to out of training set data, also increases. Adjoint-matching is a promising method of training neural network surrogates in situations where adjoint information is available. 

Future work will consider methods of incorporating derivative information to learn the dynamics of higher-dimensional systems, such as atmospheric and ocean models, and use these surrogates to accelerate 4D-Var data assimilation. We will combine traditional dimensionality reduction techniques such as POD-DEIM, as recently explored in \cite{DINO}, with the application of the adjoint-matching methodology,  to train surrogates based on more complex neural network architectures.

\section{Data Availability}
\label{sec-data-availability}
The datasets generated during and/or analysed during the current study are available from the corresponding author on reasonable request.
\bibstyle{sn-mathphys}
\bibliography{biblio,sandu}



\end{document}